\documentclass[10pt,twocolumn,letterpaper]{article}

\usepackage{cvpr}
\usepackage{times}
\usepackage{epsfig}
\usepackage{graphicx}
\usepackage{amsmath}
\usepackage{amssymb}
\usepackage{color}
\usepackage{booktabs,threeparttable}
\usepackage{colortbl}
\usepackage[export]{adjustbox}
\usepackage{caption}

\usepackage[pagebackref=true,breaklinks=true,letterpaper=true,colorlinks,bookmarks=false]{hyperref}

\cvprfinalcopy 

\def\cvprPaperID{} 

\begin{document}
	
		\title{YOLOv7: Trainable bag-of-freebies sets new state-of-the-art for real-time object detectors}
		
		\author{
			Chien-Yao Wang$^{1}$, Alexey Bochkovskiy, and Hong-Yuan Mark Liao$^{1}$ \\
			$^{1}$Institute of Information Science, Academia Sinica, Taiwan \\	
			{\tt\small kinyiu@iis.sinica.edu.tw, alexeyab84@gmail.com, and liao@iis.sinica.edu.tw}
			\vspace{-4mm}
		}
		
		\maketitle
		
		\begin{abstract}
			
			
			\vspace{-2mm}
			YOLOv7 surpasses all known object detectors in both speed and accuracy in the range from 5 FPS to 160 FPS and has the highest accuracy 56.8\% AP among all known real-time object detectors with 30 FPS or higher on GPU V100. YOLOv7-E6 object detector (56 FPS V100, 55.9\% AP) outperforms both transformer-based detector SWIN-L Cascade-Mask R-CNN (9.2 FPS A100, 53.9\% AP) by 509\% in speed and 2\% in accuracy, and convolutional-based detector ConvNeXt-XL Cascade-Mask R-CNN (8.6 FPS A100, 55.2\% AP) by 551\% in speed and 0.7\% AP in accuracy, as well as YOLOv7 outperforms: YOLOR, YOLOX, Scaled-YOLOv4, YOLOv5, DETR, Deformable DETR, DINO-5scale-R50, ViT-Adapter-B and many other object detectors in speed and accuracy. Moreover, we train YOLOv7 only on MS COCO dataset from scratch without using any other datasets or pre-trained weights. Source code is released in \url{https://github.com/WongKinYiu/yolov7}.
			\vspace{-8mm}
		\end{abstract}
		
		\section{Introduction}
		\label{sec:intr}
				
		Real-time object detection is a very important topic in computer vision, as it is often a necessary component in computer vision systems. For example, multi-object tracking \cite{zhang2021fairmot, zhang2021bytetrack}, autonomous driving \cite{li2019gs3d, feng2020deep}, robotics \cite{karaoguz2019object, paul2021object}, medical image analysis \cite{jaeger2020retina, li2019clu}, etc. The computing devices that execute real-time object detection is usually some mobile CPU or GPU, as well as various neural processing units (NPU) developed by major manufacturers. For example, the Apple neural engine (Apple), the neural compute stick (Intel), Jetson AI edge devices (Nvidia), the edge TPU (Google), the neural processing engine (Qualcomm), the AI processing unit (MediaTek), and the AI SoCs (Kneron), are all NPUs. Some of the above mentioned edge devices focus on speeding up different operations such as vanilla convolution, depth-wise convolution, or MLP operations. In this paper, the real-time object detector we proposed mainly hopes that it can support both mobile GPU and GPU devices from the edge to the cloud.
		
		\begin{figure}[t]
			\begin{center}
				\includegraphics[width=1.\linewidth]{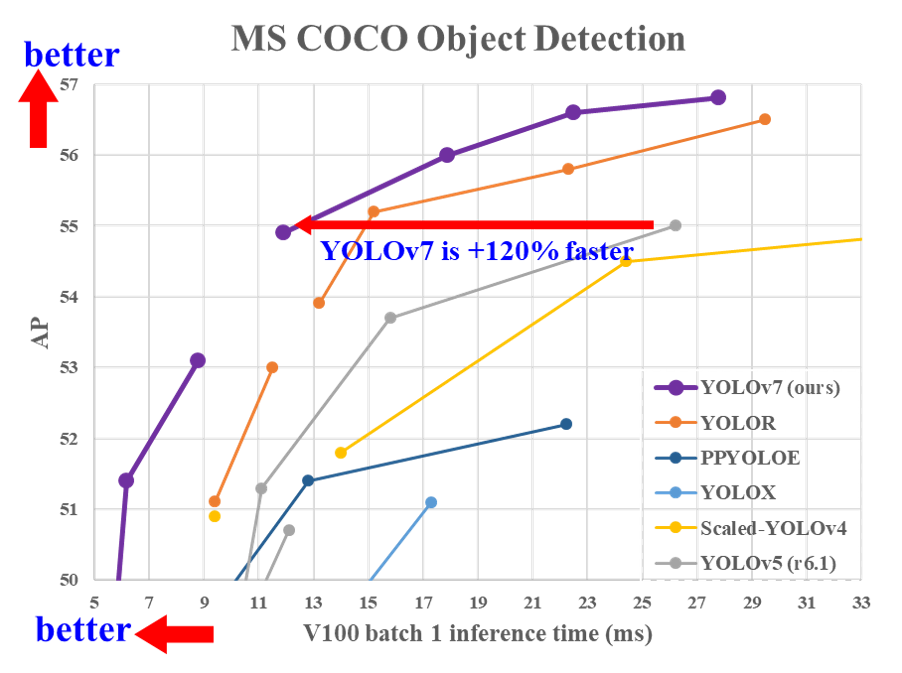}
			\end{center}
			\caption{Comparison with other real-time object detectors, our proposed methods achieve state-of-the-arts performance.}
			\label{fig:sota}
			\vspace{-2mm}
		\end{figure}
				
		In recent years, the real-time object detector is still developed for different edge device. For example, the development of MCUNet \cite{lin2020mcunet, lin2021memory} and NanoDet \cite{lyu2021nano} focused on producing low-power single-chip and improving the inference speed on edge CPU. As for methods such as YOLOX \cite{ge2021yolox} and YOLOR \cite{wang2021you}, they focus on improving the inference speed of various GPUs. More recently, the development of real-time object detector has focused on the design of efficient architecture. As for real-time object detectors that can be used on CPU \cite{lyu2021nano, yu2021pp, xiong2021mobiledets, wu2021fbnetv5}, their design is mostly based on MobileNet \cite{howard2017mobilenets, sandler2018mobilenetv2, howard2019searching}, ShuffleNet \cite{zhang2018shufflenet, ma2018shufflenetv2}, or GhostNet \cite{han2020ghostnet}. Another mainstream real-time object detectors are developed for GPU \cite{wang2021you, ge2021yolox, zhou2019objects}, they mostly use ResNet \cite{he2016deep}, DarkNet \cite{redmon2018yolov3}, or DLA \cite{yu2018deep}, and then use the CSPNet \cite{wang2020cspnet} strategy to optimize the architecture. The development direction of the proposed methods in this paper are different from that of the current mainstream real-time object detectors. In addition to architecture optimization, our proposed methods will focus on the optimization of the training process. Our focus will be on some optimized modules and optimization methods which may strengthen the training cost for improving the accuracy of object detection, but without increasing the inference cost. We call the proposed modules and optimization methods trainable bag-of-freebies.
	
		\newpage
		
		Recently, model re-parameterization \cite{ding2021repvgg, ding2021diverse, hu2022online} and dynamic label assignment \cite{ge2021ota, feng2021tood, li2022dual} have become important topics in network training and object detection. Mainly after the above new concepts are proposed, the training of object detector evolves many new issues. In this paper, we will present some of the new issues we have discovered and devise effective methods to address them. For model re-parameterization, we analyze the model re-parameterization strategies applicable to layers in different networks with the concept of gradient propagation path, and propose planned re-parameterized model. In addition, when we discover that with dynamic label assignment technology, the training of model with multiple output layers will generate new issues. That is: ``How to assign dynamic targets for the outputs of different branches?'' For this problem, we propose a new label assignment method called coarse-to-fine lead guided label assignment.
		
		The contributions of this paper are summarized as follows: (1) we design several trainable bag-of-freebies methods, so that real-time object detection can greatly improve the detection accuracy without increasing the inference cost; (2) for the evolution of object detection methods, we found two new issues, namely how re-parameterized module replaces original module, and how dynamic label assignment strategy deals with assignment to different output layers. In addition, we also propose methods to address the difficulties arising from these issues; (3) we propose ``extend'' and ``compound scaling'' methods for the real-time object detector that can effectively utilize parameters and computation; and (4) the method we proposed can effectively reduce about 40\% parameters and 50\% computation of state-of-the-art real-time object detector, and has faster inference speed and higher detection accuracy.
		
		\section{Related work}
		\label{sec:relw}

		\subsection{Real-time object detectors}
		
		Currently state-of-the-art real-time object detectors are mainly based on YOLO \cite{redmon2016you, redmon2017yolo9000, redmon2018yolov3} and FCOS \cite{tian2019fcos, tian2022fcos}, which are \cite{bochkovskiy2020yolov4, wang2021scaled, wang2021you, ge2021yolox, lyu2021nano, xu2022pp, glenn2022yolov5}. Being able to become a state-of-the-art real-time object detector usually requires the following characteristics: (1) a faster and stronger network architecture; (2) a more effective feature integration method \cite{ghiasi2019fpn, zhou2019objects, kirillov2019panoptic, tan2020efficientdet, qiao2021detectors, hu2021a2, dai2021dynamic, li2022exploring}; (3) a more accurate detection method \cite{tian2019fcos, tian2022fcos, sun2021sparse}; (4) a more robust loss function \cite{zhou2019iou, rezatofighi2019generalized, chen2020ap, oksuz2020ranking, zheng2020distance, oksuz2021rank}; (5) a more efficient label assignment method \cite{zhu2020autoassign, ge2021ota, feng2021tood, wang2021end, li2022dual}; and (6) a more efficient training method. In this paper, we do not intend to explore self-supervised learning or knowledge distillation methods that require additional data or large model. Instead, we will design new trainable bag-of-freebies method for the issues derived from the state-of-the-art methods associated with (4), (5), and (6) mentioned above.
		
		\subsection{Model re-parameterization}
		
		Model re-parametrization techniques \cite{szegedy2016rethinking, huang2017snapshot, tarvainen2017mean, garipov2018loss, izmailov2018averaging, ding2019acnet, cao2020ensemble, guo2020expandnets, ding2021repvgg, ding2021diverse, ding2022re, hu2022online, ding2022scaling, vasu2022improved} merge multiple computational modules into one at inference stage. The model re-parameterization technique can be regarded as an ensemble technique, and we can divide it into two categories, i.e., module-level ensemble and model-level ensemble. There are two common practices for model-level re-parameterization to obtain the final inference model. One is to train multiple identical models with different training data, and then average the weights of multiple trained models.  The other is to perform a weighted average of the weights of models at different iteration number. Module-level re-parameterization is a more popular research issue recently. This type of method splits a module into multiple identical or different module branches during training and integrates multiple branched modules into a completely equivalent module during inference. However, not all proposed re-parameterized module can be perfectly applied to different architectures. With this in mind, we have developed new re-parameterization module and designed related application strategies for various architectures.
		
		\subsection{Model scaling}
		
		Model scaling \cite{tan2019efficientnet, radosavovic2020designing, tan2020efficientdet, tan2021efficientnetv2, dollar2021fast, du2021simple, bello2021revisiting, liu2022swin} is a way to scale up or down an already designed model and make it fit in different computing devices. The model scaling method usually uses different scaling factors, such as resolution (size of input image), depth (number of layer), width (number of channel), and stage (number of feature pyramid), so as to achieve a good trade-off for the amount of network parameters, computation, inference speed, and accuracy. Network architecture search (NAS) is one of the commonly used model scaling methods. NAS can automatically search for suitable scaling factors from search space without defining too complicated rules. The disadvantage of NAS is that it requires very expensive computation to complete the search for model scaling factors. In \cite{dollar2021fast}, the researcher analyzes the relationship between scaling factors and the amount of parameters and operations, trying to directly estimate some rules, and thereby obtain the scaling factors required by model scaling. Checking the literature, we found that almost all model scaling methods analyze individual scaling factor independently, and even the methods in the compound scaling category also optimized scaling factor independently. The reason for this is because most popular NAS architectures deal with scaling factors that are not very correlated. We observed that all concatenation-based models, such as DenseNet \cite{huang2017densely} or VoVNet \cite{lee2019energy}, will change the input width of some layers when the depth of such models is scaled. Since the proposed architecture is concatenation-based, we have to design a new compound scaling method for this model.
		
		\begin{figure*}[t]
			\begin{center}
				\includegraphics[width=1.\linewidth]{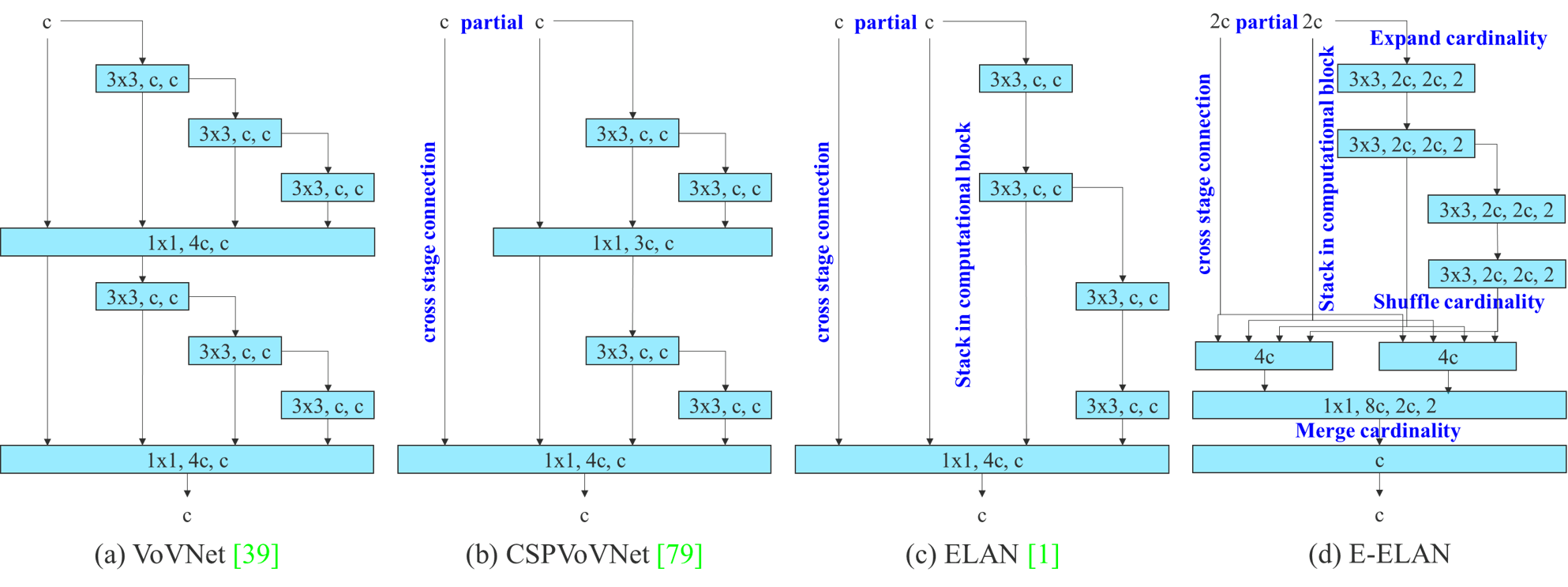}
			\end{center}
			\caption{Extended efficient layer aggregation networks.  The proposed extended ELAN (E-ELAN) does not change the gradient transmission path of the original architecture at all, but use group convolution to increase the cardinality of the added features, and combine the features of different groups in a shuffle and merge cardinality manner.  This way of operation can enhance the features learned by different feature maps and improve the use of parameters and calculations.}
			\label{fig:elan}
			\vspace{-2mm}
		\end{figure*}

		\newpage
		
		\section{Architecture}

		\subsection{Extended efficient layer aggregation networks}
		
		In most of the literature on designing the efficient architectures, the main considerations are no more than the number of parameters, the amount of computation, and the computational density. Starting from the characteristics of memory access cost, Ma \etal \cite{ma2018shufflenetv2} also analyzed the influence of the input/output channel ratio, the number of branches of the architecture, and the element-wise operation on the network inference speed. Doll{\'a}r \etal \cite{dollar2021fast} additionally considered activation when performing model scaling, that is, to put more consideration on the number of elements in the output tensors of convolutional layers. The design of CSPVoVNet \cite{wang2021scaled} in Figure \ref{fig:elan} (b) is a variation of VoVNet \cite{lee2019energy}. In addition to considering the aforementioned basic designing concerns, the architecture of CSPVoVNet \cite{wang2021scaled} also analyzes the gradient path, in order to enable the weights of different layers to learn more diverse features. The gradient analysis approach described above makes inferences faster and more accurate. ELAN \cite{wang2022designing} in Figure \ref{fig:elan} (c) considers the following design strategy -- ``How to design an efficient network?.'' They came out with a conclusion: By controlling the shortest longest gradient path, a deeper network can learn and converge effectively. In this paper, we propose Extended-ELAN (E-ELAN) based on ELAN and its main architecture is shown in Figure \ref{fig:elan} (d).
		
		Regardless of the gradient path length and the stacking number of computational blocks in large-scale ELAN, it has reached a stable state. If more computational blocks are stacked unlimitedly, this stable state may be destroyed, and the parameter utilization rate will decrease. The proposed E-ELAN uses expand, shuffle, merge cardinality to achieve the ability to continuously enhance the learning ability of the network without destroying the original gradient path. In terms of architecture, E-ELAN only changes the architecture in computational block, while the architecture of transition layer is completely unchanged. Our strategy is to use group convolution to expand the channel and cardinality of computational blocks. We will apply the same group parameter and channel multiplier to all the computational blocks of a computational layer. Then, the feature map calculated by each computational block will be shuffled into $g$ groups according to the set group parameter $g$, and then concatenate them together. At this time, the number of channels in each group of feature map will be the same as the number of channels in the original architecture. Finally, we add $g$ groups of feature maps to perform merge cardinality. In addition to maintaining the original ELAN design architecture, E-ELAN can also guide different groups of computational blocks to learn more diverse features.
		
		\begin{figure*}[t]
			\begin{center}
				\includegraphics[width=1.0\linewidth]{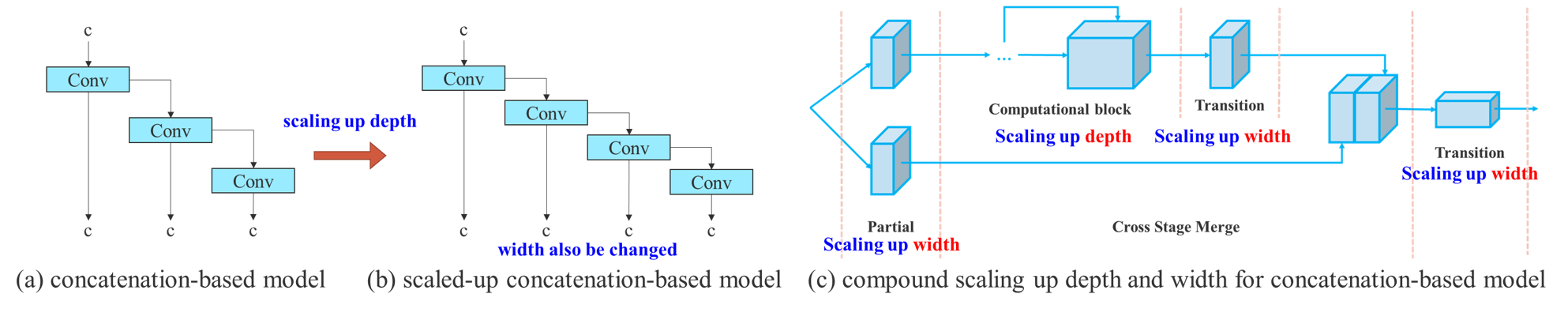}
			\end{center}
			\caption{Model scaling for concatenation-based models.  From (a) to (b), we observe that when depth scaling is performed on concatenation-based models, the output width of a computational block also increases.  This phenomenon will cause the input width of the subsequent transmission layer to increase.  Therefore, we propose (c), that is, when performing model scaling on concatenation-based models, only the depth in a computational block needs to be scaled, and the remaining of transmission layer is performed with corresponding width scaling.}
			\label{fig:cosc}
			\vspace{-2mm}
		\end{figure*}
		
		
		\subsection{Model scaling for concatenation-based models}
		
		The main purpose of model scaling is to adjust some attributes of the model and generate models of different scales to meet the needs of different inference speeds. For example the scaling model of EfficientNet \cite{tan2019efficientnet} considers the width, depth, and resolution. As for the scaled-YOLOv4 \cite{wang2021scaled}, its scaling model is to adjust the number of stages. In \cite{dollar2021fast}, Doll{\'a}r \etal analyzed the influence of vanilla convolution and group convolution on the amount of parameter and computation when performing width and depth scaling, and used this to design the corresponding model scaling method. The above methods are mainly used in architectures such as PlainNet or ResNet. When these architectures are in executing scaling up or scaling down, the in-degree and out-degree of each layer will not change, so we can independently analyze the impact of each scaling factor on the amount of parameters and computation. However, if these methods are applied to the concatenation-based architecture, we will find that when scaling up or scaling down is performed on depth, the in-degree of a translation layer which is immediately after a concatenation-based computational block will decrease or increase, as shown in Figure \ref{table:scale} (a) and (b).
		
		It can be inferred from the above phenomenon that we cannot analyze different scaling factors separately for a concatenation-based model but must be considered together. Take scaling-up depth as an example, such an action will cause a ratio change between the input channel and output channel of a transition layer, which may lead to a decrease in the hardware usage of the model. Therefore, we must propose the corresponding compound model scaling method for a concatenation-based model. When we scale the depth factor of a computational block, we must also calculate the change of the output channel of that block. Then, we will perform width factor scaling with the same amount of change on the transition layers, and the result is shown in Figure \ref{table:scale} (c). Our proposed compound scaling method can maintain the properties that the model had at the initial design and maintains the optimal structure.
		
		
		\section{Trainable bag-of-freebies}
		
		\subsection{Planned re-parameterized convolution}
		
		Although RepConv \cite{ding2021repvgg} has achieved excellent performance on the VGG \cite{simonyan2014very}, when we directly apply it to ResNet \cite{he2016deep} and DenseNet \cite{huang2017densely} and other architectures, its accuracy will be significantly reduced. We use gradient flow propagation paths to analyze how re-parameterized convolution should be combined with different network. We also designed planned re-parameterized convolution accordingly.
		
		\begin{figure}[h]
			\begin{center}
				\includegraphics[width=1.0\linewidth]{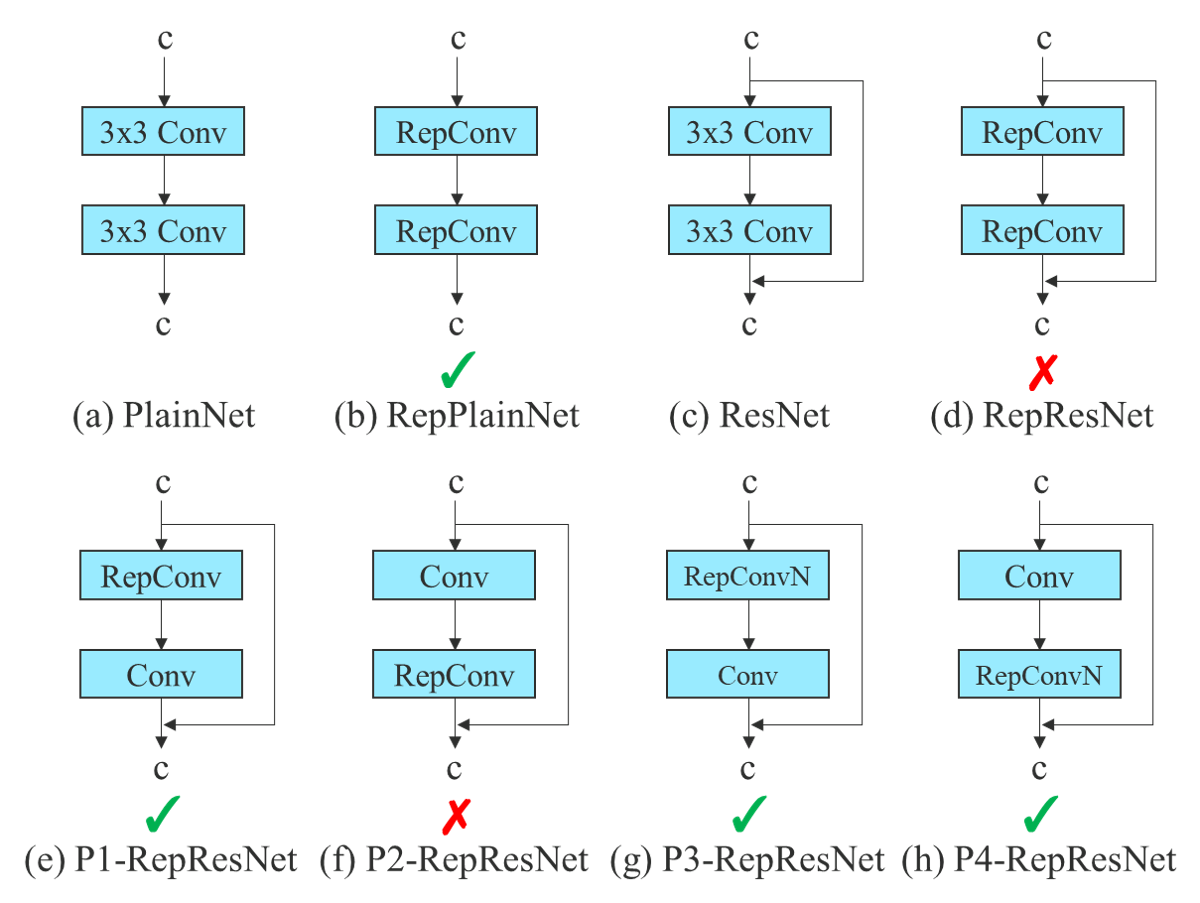}
			\end{center}
			\caption{Planned re-parameterized model.  In the proposed planned re-parameterized model, we found that a layer with residual or concatenation connections, its RepConv should not have identity connection.  Under these circumstances, it can be replaced by RepConvN that contains no identity connections.}
			\label{fig:prep}
			\vspace{-2mm}
		\end{figure}
		
		RepConv actually combines $3 \times 3$ convolution,  $1 \times 1$ convolution, and identity connection in one convolutional layer. After analyzing the combination and corresponding performance of RepConv and different architectures, we find that the identity connection in RepConv destroys the residual in ResNet and the concatenation in DenseNet, which provides more diversity of gradients for different feature maps. For the above reasons, we use RepConv without identity connection (RepConvN) to design the architecture of planned re-parameterized convolution. In our thinking, when a convolutional layer with residual or concatenation is replaced by re-parameterized convolution, there should be no identity connection. Figure \ref{fig:prep} shows an example of our designed ``planned re-parameterized convolution'' used in PlainNet and ResNet. As for the complete planned re-parameterized convolution experiment in residual-based model and concatenation-based model, it will be presented in the ablation study session.
		
		\begin{figure*}[h]
			\begin{center}
				\includegraphics[width=1.0\linewidth]{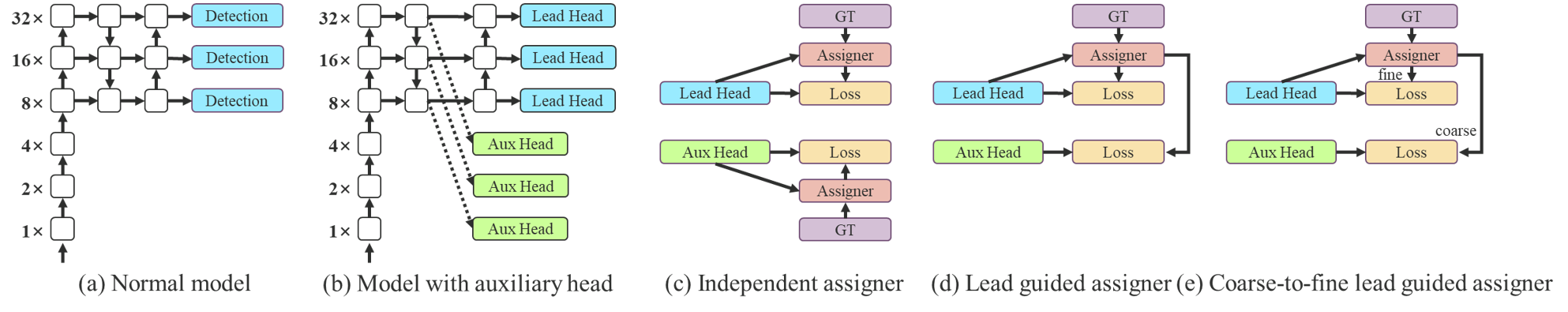}
			\end{center}
			\caption{Coarse for auxiliary and fine for lead head label assigner.  Compare with normal model (a), the schema in (b) has auxiliary head. Different from the usual independent label assigner (c), we propose (d) lead head guided label assigner and (e) coarse-to-fine lead head guided label assigner.  The proposed label assigner is optimized by lead head prediction and the ground truth to get the labels of training lead head and auxiliary head at the same time.  The detailed coarse-to-fine implementation method and constraint design details will be elaborated in Apendix.}
			\label{fig:ctof}
		\end{figure*}
		

		\subsection{Coarse for auxiliary and fine for lead loss}
		
		Deep supervision \cite{lee2015deeply} is a technique that is often used in training deep networks.  Its main concept is to add extra auxiliary head in the middle layers of the network, and the shallow network weights with assistant loss as the guide.  Even for architectures such as ResNet \cite{he2016deep} and DenseNet \cite{huang2017densely} which usually converge well, deep supervision \cite{szegedy2015going, zhou2018unetpp, shen2019object, liang2021cbnetv2, wang2021end, roh2021sparse, yang20213d, liu2021yolostereo3d} can still significantly improve the performance of the model on many tasks.  Figure \ref{fig:ctof} (a) and (b) show, respectively, the object detector architecture ``without'' and ``with'' deep supervision.  In this paper, we call the head responsible for the final output as the lead head, and the head used to assist training is called auxiliary head.
		
		Next we want to discuss the issue of label assignment.  In the past, in the training of deep network, label assignment usually refers directly to the ground truth and generate hard label according to the given rules.  However, in recent years, if we take object detection as an example, researchers often use the quality and distribution of prediction output by the network, and then consider together with the ground truth to use some calculation and optimization methods to generate a reliable soft label \cite{redmon2016you, choi2019gaussian, kim2020probabilistic, zhu2020autoassign, zhang2020bridging, li2020generalized, li2021generalized, zhang2021varifocalnet, ge2021ota, feng2021tood, li2022dual}. For example, YOLO \cite{redmon2016you} use IoU of prediction of bounding box regression and ground truth as the soft label of objectness.  In this paper, we call the mechanism that considers the network prediction results together with the ground truth and then assigns soft labels as ``label assigner.''
		
		Deep supervision needs to be trained on the target objectives regardless of the circumstances of auxiliary head or lead head.  During the development of soft label assigner related techniques, we accidentally discovered a new derivative issue, i.e., ``How to assign soft label to auxiliary head and lead head ?''  To the best of our knowledge, the relevant literature has not explored this issue so far.  The results of the most popular method at present is as shown in Figure \ref{fig:ctof} (c), which is to separate auxiliary head and lead head, and then use their own prediction results and the ground truth to execute label assignment.  The method proposed in this paper is a new label assignment method that guides both auxiliary head and lead head by the lead head prediction.  In other words, we use lead head prediction as guidance to generate coarse-to-fine hierarchical labels, which are used for auxiliary head and lead head learning, respectively.  The two proposed deep supervision label assignment strategies are shown in Figure \ref{fig:ctof} (d) and (e), respectively.
				
		\textbf{Lead head guided label assigner} is mainly calculated based on the prediction result of the lead head and the ground truth, and generate soft label through the optimization process.  This set of soft labels will be used as the target training model for both auxiliary head and lead head.  The reason to do this is because lead head has a relatively strong learning capability, so the soft label generated from it should be more representative of the distribution and correlation between the source data and the target.  Furthermore, we can view such learning as a kind of generalized residual learning.  By letting the shallower auxiliary head directly learn the information that lead head has learned, lead head will be more able to focus on learning residual information that has not yet been learned.
		
		\begin{table*}[t]
			\centering
			\begin{threeparttable}[t]
				\footnotesize
				\caption{Comparison of baseline object detectors.}
				\label{table:base}
				\begin{tabular}{lccccccccc}
					\toprule
					\textbf{Model} & \textbf{\#Param.} & \textbf{FLOPs} & \textbf{Size} & \textbf{AP$^{val}$} & \textbf{AP$^{val}_{50}$} & \textbf{AP$^{val}_{75}$} & \textbf{AP$^{val}_S$} & \textbf{AP$^{val}_M$} & \textbf{AP$^{val}_L$} \\	
					\midrule
					\textbf{YOLOv4 \cite{bochkovskiy2020yolov4}} & 64.4M & 142.8G & 640 & 49.7\% & 68.2\% & 54.3\% & 32.9\% & 54.8\% & 63.7\% \\
					\textbf{YOLOR-u5 (r6.1) \cite{wang2021you}} & 46.5M & 109.1G & 640 & 50.2\% & 68.7\% & 54.6\% & 33.2\% & 55.5\% & 63.7\% \\
					\textbf{YOLOv4-CSP \cite{wang2021scaled}} & 52.9M & 120.4G & 640 & 50.3\% & 68.6\% & 54.9\% & 34.2\% & 55.6\% & 65.1\% \\
					\textbf{YOLOR-CSP \cite{wang2021you}} & 52.9M & 120.4G & 640 & 50.8\% & 69.5\% & 55.3\% & 33.7\% & 56.0\% & 65.4\% \\
					\textbf{YOLOv7} & 36.9M & 104.7G & 640 & \textbf{51.2\%} & \textbf{69.7\%} & \textbf{55.5\%} & \textbf{35.2\%} & \textbf{56.0\%} & \textbf{66.7\%} \\
					improvement & \textcolor{green}{-43\%} & \textcolor{green}{-15\%} & - & \textcolor{green}{+0.4} & \textcolor{green}{+0.2} & \textcolor{green}{+0.2} & \textcolor{green}{+1.5} & \textcolor{cyan}{=} & \textcolor{green}{+1.3} \\
					\midrule
					\textbf{YOLOR-CSP-X \cite{wang2021you}} & 96.9M & 226.8G & 640 & 52.7\% & \textbf{71.3\%} & 57.4\% & 36.3\% & 57.5\% & 68.3\% \\
					\textbf{YOLOv7-X} & 71.3M & 189.9G & 640 & \textbf{52.9\%} & 71.1\% & \textbf{57.5\%} & \textbf{36.9\%} & \textbf{57.7\%} & \textbf{68.6\%} \\
					improvement & \textcolor{green}{-36\%} & \textcolor{green}{-19\%} & - & \textcolor{green}{+0.2} & \textcolor{red}{-0.2} & \textcolor{green}{+0.1} & \textcolor{green}{+0.6} & \textcolor{green}{+0.2} & \textcolor{green}{+0.3} \\
					\midrule
					\textbf{YOLOv4-tiny \cite{wang2021scaled}} & 6.1 & 6.9 & 416 & 24.9\% & 42.1\% & 25.7\% & 8.7\% & 28.4\% & 39.2\% \\
					\textbf{YOLOv7-tiny} & 6.2 & 5.8 & 416 & \textbf{35.2\%} & \textbf{52.8\%} & \textbf{37.3\%} & \textbf{15.7\%} & \textbf{38.0\%} & \textbf{53.4\%} \\
					improvement & \textcolor{red}{+2\%} & \textcolor{green}{-19\%} & - & \textcolor{green}{+10.3} & \textcolor{green}{+10.7} & \textcolor{green}{+11.6} & \textcolor{green}{+7.0} & \textcolor{green}{+9.6} & \textcolor{green}{+14.2} \\
					\midrule
					\textbf{YOLOv4-tiny-3l \cite{wang2021scaled}} & 8.7 & 5.2 & 320 & 30.8\% & 47.3\% & 32.2\% & \textbf{10.9\%} & 31.9\% & 51.5\% \\
					\textbf{YOLOv7-tiny} & 6.2 & 3.5 & 320 & \textbf{30.8\%} & \textbf{47.3\%} & \textbf{32.2\%} & 10.0\% & \textbf{31.9\%} & \textbf{52.2\%} \\
					improvement & \textcolor{green}{-39\%} & \textcolor{green}{-49\%} & - & \textcolor{cyan}{=} & \textcolor{cyan}{=} & \textcolor{cyan}{=} & \textcolor{red}{-0.9} & \textcolor{cyan}{=} & \textcolor{green}{+0.7} \\
					\midrule
					\textbf{YOLOR-E6 \cite{wang2021you}} & 115.8M & 683.2G & 1280 & 55.7\% & 73.2\% & 60.7\% & 40.1\% & \textbf{60.4\%} & 69.2\% \\
					\textbf{YOLOv7-E6} & 97.2M & 515.2G & 1280 & \textbf{55.9\%} & \textbf{73.5\%} & \textbf{61.1\%} & \textbf{40.6\%} & 60.3\% & \textbf{70.0\%} \\
					improvement & \textcolor{green}{-19\%} & \textcolor{green}{-33\%} & - & \textcolor{green}{+0.2} & \textcolor{green}{+0.3} & \textcolor{green}{+0.4} & \textcolor{green}{+0.5} & \textcolor{red}{-0.1} & \textcolor{green}{+0.8} \\
					\midrule
					\textbf{YOLOR-D6 \cite{wang2021you}} & 151.7M & 935.6G & 1280 & 56.1\% & 73.9\% & 61.2\% & \textbf{42.4\%} & 60.5\% & 69.9\% \\
					\textbf{YOLOv7-D6} & 154.7M & 806.8G & 1280 & 56.3\% & 73.8\% & 61.4\% & 41.3\% & 60.6\% & 70.1\% \\
					\textbf{YOLOv7-E6E} & 151.7M & 843.2G & 1280 & \textbf{56.8\%} & \textbf{74.4\%} & \textbf{62.1\%} & 40.8\% & \textbf{62.1\%} & \textbf{70.6\%} \\
					improvement & \textcolor{cyan}{=} & \textcolor{green}{-11\%} & - & \textcolor{green}{+0.7} & \textcolor{green}{+0.5} & \textcolor{green}{+0.9} & \textcolor{red}{-1.6} & \textcolor{green}{+1.6} & \textcolor{green}{+0.7} \\
					\bottomrule
				\end{tabular}
			\end{threeparttable}
		\vspace{-2mm}
		\end{table*}
		
		\textbf{Coarse-to-fine lead head guided label assigner} also used the predicted result of the lead head and the ground truth  to generate soft label.  However, in the process we generate two different sets of soft label, i.e., coarse label and fine label, where fine label is the same as the soft label generated by lead head guided label assigner, and coarse label is generated by allowing more grids to be treated as positive target by relaxing the constraints of the positive sample assignment process.  The reason for this is that the learning ability of an auxiliary head is not as strong as that of a lead head, and in order to avoid losing the information that needs to be learned, we will focus on optimizing the recall of auxiliary head in the object detection task.  As for the output of lead head, we can filter the high precision results from the high recall results as the final output.  However, we must note that if the additional weight of coarse label is close to that of fine label, it may produce bad prior at final prediction.  Therefore, in order to make those extra coarse positive grids have less impact, we put restrictions in the decoder, so that the extra coarse positive grids cannot produce soft label perfectly.  The mechanism mentioned above allows the importance of fine label and coarse label to be dynamically adjusted during the learning process, and makes the optimizable upper bound of fine label always higher than coarse label.	
		
		\subsection{Other trainable bag-of-freebies}
		
		In this section we will list some trainable bag-of-freebies.  These freebies are some of the tricks we used in training, but the original concepts were not proposed by us.  The training details of these freebies will be elaborated in the Appendix, including (1) Batch normalization in conv-bn-activation topology: This part mainly connects batch normalization layer directly to convolutional layer.  The purpose of this is to integrate the mean and variance of batch normalization into the bias and weight of convolutional layer at the inference stage. (2) Implicit knowledge in YOLOR \cite{wang2021you} combined with convolution feature map in addition and multiplication manner: Implicit knowledge in YOLOR can be simplified to a vector by pre-computing at the inference stage.  This vector can be combined with the bias and weight of the previous or subsequent convolutional layer. (3) EMA model: EMA is a technique used in mean teacher \cite{tarvainen2017mean}, and in our system we use EMA model purely as the final inference model.
		
		\section{Experiments}

		\subsection{Experimental setup}
		
		We use Microsoft COCO dataset to conduct experiments and validate our object detection method. All our experiments did not use pre-trained models. That is, all models were trained from scratch. During the development process, we used train 2017 set for training, and then used val 2017 set for verification and choosing hyperparameters. Finally, we show the performance of object detection on the test 2017 set and compare it with the state-of-the-art object detection algorithms. Detailed training parameter settings are described in Appendix.
		
		We designed basic model for edge GPU, normal GPU, and cloud GPU, and they are respectively called YOLOv7-tiny, YOLOv7, and YOLOv7-W6. At the same time, we also use basic model for model scaling for different service requirements and get different types of models. For YOLOv7, we do stack scaling on neck, and use the proposed compound scaling method to perform scaling-up of the depth and width of the entire model, and use this to obtain YOLOv7-X. As for YOLOv7-W6, we use the newly proposed compound scaling method to obtain YOLOv7-E6 and YOLOv7-D6. In addition, we use the proposed E-ELAN for YOLOv7-E6, and thereby complete YOLOv7-E6E. Since YOLOv7-tiny is an edge GPU-oriented architecture, it will use leaky ReLU as activation function. As for other models we use SiLU as activation function. We will describe the scaling factor of each model in detail in Appendix.

		
		\begin{table*}[t]
			\centering
			\begin{threeparttable}[t]
				\footnotesize
				\caption{Comparison of state-of-the-art real-time object detectors.}
				\label{table:sota}
				\begin{tabular}{l|c|c|c|c|c|ccccc}
					\toprule
					\textbf{Model} & \textbf{\#Param.} & \textbf{FLOPs} & \textbf{Size} & \textbf{FPS} & \textbf{AP$^{test}$} / \textbf{AP$^{val}$} & \textbf{AP$^{test}_{50}$} & \textbf{AP$^{test}_{75}$} & \textbf{AP$^{test}_S$} & \textbf{AP$^{test}_M$} & \textbf{AP$^{test}_L$} \\	
					\midrule
					\textbf{YOLOX-S \cite{ge2021yolox}} & 9.0M & 26.8G & 640 & 102 & 40.5\% / 40.5\% & - & - & - & - & - \\
					\textbf{YOLOX-M \cite{ge2021yolox}} & 25.3M & 73.8G & 640 & 81 & 47.2\% / 46.9\% & - & - & - & - & - \\
					\textbf{YOLOX-L \cite{ge2021yolox}} & 54.2M & 155.6G & 640 & 69 & 50.1\% / 49.7\% & - & - & - & - & - \\
					\textbf{YOLOX-X \cite{ge2021yolox}} & 99.1M & 281.9G & 640 & 58 & 51.5\% / 51.1\% & - & - & - & - & - \\
					\midrule
					\textbf{PPYOLOE-S \cite{xu2022pp}} & 7.9M & 17.4G & 640 & 208 & 43.1\% / 42.7\% & 60.5\% & 46.6\% & 23.2\% & 46.4\% & 56.9\% \\
					\textbf{PPYOLOE-M \cite{xu2022pp}} & 23.4M & 49.9G & 640 & 123 & 48.9\% / 48.6\% & 66.5\% & 53.0\% & 28.6\% & 52.9\% & 63.8\% \\
					\textbf{PPYOLOE-L \cite{xu2022pp}} & 52.2M & 110.1G & 640 & 78 & 51.4\% / 50.9\% & 68.9\% & 55.6\% & 31.4\% & 55.3\% & 66.1\% \\
					\textbf{PPYOLOE-X \cite{xu2022pp}} & 98.4M & 206.6G & 640 & 45 & 52.2\% / 51.9\% & 69.9\% & 56.5\% & 33.3\% & 56.3\% & 66.4\% \\
					\midrule
					\textbf{YOLOv5-N (r6.1) \cite{glenn2022yolov5}} & 1.9M & 4.5G & 640 & 159 & - / 28.0\% & - & - & - & - & - \\
					\textbf{YOLOv5-S (r6.1) \cite{glenn2022yolov5}} & 7.2M & 16.5G & 640 & 156 & - / 37.4\% & - & - & - & - & - \\
					\textbf{YOLOv5-M (r6.1) \cite{glenn2022yolov5}} & 21.2M & 49.0G & 640 & 122 & - / 45.4\% & - & - & - & - & - \\
					\textbf{YOLOv5-L (r6.1) \cite{glenn2022yolov5}} & 46.5M & 109.1G & 640 & 99 & - / 49.0\% & - & - & - & - & - \\
					\textbf{YOLOv5-X (r6.1) \cite{glenn2022yolov5}} & 86.7M & 205.7G & 640 & 83 & - / 50.7\% & - & - & - & - & - \\
					\midrule
					\textbf{YOLOR-CSP \cite{wang2021you}} & 52.9M & 120.4G & 640 & 106 & 51.1\% / 50.8\% & 69.6\% & 55.7\% & 31.7\% & 55.3\% & 64.7\% \\
					\textbf{YOLOR-CSP-X \cite{wang2021you}} & 96.9M & 226.8G & 640 & 87 & 53.0\% / 52.7\% & 71.4\% & 57.9\% & 33.7\% & 57.1\% & 66.8\% \\
					\midrule
					\textbf{YOLOv7-tiny-SiLU} & 6.2M & 13.8G & 640 & 286 & 38.7\% / 38.7\% & 56.7\% & 41.7\% & 18.8\% & 42.4\% & 51.9\% \\
					\textbf{YOLOv7} & 36.9M & 104.7G & 640 & 161 & 51.4\% / 51.2\% & 69.7\% & 55.9\% & 31.8\% & 55.5\% & 65.0\% \\
					\textbf{YOLOv7-X} & 71.3M & 189.9G & 640 & 114 & 53.1\% / 52.9\% & 71.2\% & 57.8\% & 33.8\% & 57.1\% & 67.4\% \\
					\midrule
					\midrule
					\textbf{YOLOv5-N6 (r6.1) \cite{glenn2022yolov5}} & 3.2M & 18.4G & 1280 & 123 & - / 36.0\% & - & - & - & - & - \\
					\textbf{YOLOv5-S6 (r6.1) \cite{glenn2022yolov5}} & 12.6M & 67.2G & 1280 & 122 & - / 44.8\% & - & - & - & - & - \\
					\textbf{YOLOv5-M6 (r6.1) \cite{glenn2022yolov5}} & 35.7M & 200.0G & 1280 & 90 & - / 51.3\% & - & - & - & - & - \\
					\textbf{YOLOv5-L6 (r6.1) \cite{glenn2022yolov5}} & 76.8M & 445.6G & 1280 & 63 & - / 53.7\% & - & - & - & - & - \\
					\textbf{YOLOv5-X6 (r6.1) \cite{glenn2022yolov5}} & 140.7M & 839.2G & 1280 & 38 & - / 55.0\% & - & - & - & - & - \\
					\midrule
					\textbf{YOLOR-P6 \cite{wang2021you}} & 37.2M & 325.6G & 1280 & 76 & 53.9\% / 53.5\% & 71.4\% & 58.9\% & 36.1\% & 57.7\% & 65.6\% \\
					\textbf{YOLOR-W6 \cite{wang2021you}} & 79.8G & 453.2G & 1280 & 66 & 55.2\% / 54.8\% & 72.7\% & 60.5\% & 37.7\% & 59.1\% & 67.1\% \\
					\textbf{YOLOR-E6 \cite{wang2021you}} & 115.8M & 683.2G & 1280 & 45 & 55.8\% / 55.7\% & 73.4\% & 61.1\% & 38.4\% & 59.7\% & 67.7\% \\
					\textbf{YOLOR-D6 \cite{wang2021you}} & 151.7M & 935.6G & 1280 & 34 & 56.5\% / 56.1\% & 74.1\% & 61.9\% & 38.9\% & 60.4\% & 68.7\% \\
					\midrule
					\textbf{YOLOv7-W6} & 70.4M & 360.0G & 1280 & 84 & 54.9\% / 54.6\% & 72.6\% & 60.1\% & 37.3\% & 58.7\% & 67.1\% \\
					\textbf{YOLOv7-E6} & 97.2M & 515.2G & 1280 & 56 & 56.0\% / 55.9\% & 73.5\% & 61.2\% & 38.0\% & 59.9\% & 68.4\% \\
					\textbf{YOLOv7-D6} & 154.7M & 806.8G & 1280 & 44 & 56.6\% / 56.3\% & 74.0\% & 61.8\% & 38.8\% & 60.1\% & 69.5\% \\
					\textbf{YOLOv7-E6E} & 151.7M & 843.2G & 1280 & 36 & 56.8\% / 56.8\% & 74.4\% & 62.1\% & 39.3\% & 60.5\% & 69.0\% \\
					\bottomrule
				\end{tabular}
				\begin{tablenotes}[flushleft]
				\footnotesize
				\item[1] Our FLOPs is calaculated by rectangle input resolution like 640 $\times$ 640 or 1280 $\times$ 1280. 
				\item[2] Our inference time is estimated by using letterbox resize input image to make its long side equals to 640 or 1280. 
				\end{tablenotes}
			\end{threeparttable}
		\vspace{-4mm}
		\end{table*}
		
		\subsection{Baselines}
		
		We choose previous version of YOLO \cite{bochkovskiy2020yolov4, wang2021scaled} and state-of-the-art object detector YOLOR \cite{wang2021you} as our baselines. Table \ref{table:base} shows the comparison of our proposed YOLOv7 models and those baseline that are trained with the same settings.
		
		From the results we see that if compared with YOLOv4, YOLOv7 has 75\% less parameters, 36\% less computation, and brings 1.5\% higher AP. If compared with state-of-the-art YOLOR-CSP, YOLOv7 has 43\% fewer parameters, 15\% less computation, and 0.4\% higher AP. In the performance of tiny model, compared with YOLOv4-tiny-31, YOLOv7-tiny reduces the number of parameters by 39\% and the amount of computation by 49\%, but maintains the same AP. On the cloud GPU model, our model can still have a higher AP while reducing the number of parameters by 19\% and the amount of computation by 33\%.
		
		\newpage
		
		\subsection{Comparison with state-of-the-arts}
		
		We compare the proposed method with state-of-the-art object detectors for general GPUs and Mobile GPUs, and the results are shown in Table \ref{table:sota}. From the results in Table \ref{table:sota} we know that the proposed method has the best speed-accuracy trade-off comprehensively. If we compare YOLOv7-tiny-SiLU with YOLOv5-N (r6.1), our method is 127 fps faster and 10.7\% more accurate on AP.  In addition, YOLOv7 has 51.4\% AP at frame rate of 161 fps, while PPYOLOE-L with the same AP has only 78 fps frame rate. In terms of parameter usage, YOLOv7 is 41\% less than PPYOLOE-L. If we compare YOLOv7-X with 114 fps inference speed to YOLOv5-L (r6.1) with 99 fps inference speed, YOLOv7-X can improve AP by 3.9\%. If YOLOv7-X is compared with YOLOv5-X (r6.1) of similar scale, the inference speed of YOLOv7-X is 31 fps faster. In addition, in terms of the amount of parameters and computation, YOLOv7-X reduces 22\% of parameters and 8\% of computation compared to YOLOv5-X (r6.1), but improves AP by 2.2\%.
		
		\newpage
		
		If we compare YOLOv7 with YOLOR using the input resolution 1280, the inference speed of YOLOv7-W6 is 8 fps faster than that of YOLOR-P6, and the detection rate is also increased by 1\% AP. As for the comparison between YOLOv7-E6 and YOLOv5-X6 (r6.1), the former has 0.9\% AP gain than the latter, 45\% less parameters and 63\% less computation, and the inference speed is increased by 47\%. YOLOv7-D6 has close inference speed to YOLOR-E6, but improves AP by 0.8\%. YOLOv7-E6E has close inference speed to YOLOR-D6, but improves AP by 0.3\%.
		

		\subsection{Ablation study}

		\subsubsection{Proposed compound scaling method}
		
		Table \ref{table:scale} shows the results obtained when using different model scaling strategies for scaling up. Among them, our proposed compound scaling method is to scale up the depth of computational block by 1.5 times and the width of transition block by 1.25 times. If our method is compared with the method that only scaled up the width, our method can improve the AP by 0.5\% with less parameters and amount of computation. If our method is compared with the method that only scales up the depth, our method only needs to increase the number of parameters by 2.9\% and the amount of computation by 1.2\%, which can improve the AP by 0.2\%. It can be seen from the results of Table \ref{table:scale} that our proposed compound scaling strategy can utilize parameters and computation more efficiently.
		
		\begin{table}[h]
			\centering
			\begin{threeparttable}[h]
				\footnotesize
				\caption{Ablation study on proposed model scaling.}
				\label{table:scale}
				\setlength\tabcolsep{2.0pt}
				\begin{tabular}{lcccccc}
					\toprule
					\textbf{Model} & \textbf{\#Param.} & \textbf{FLOPs} & \textbf{Size} & \textbf{AP$^{val}$} & \textbf{AP$^{val}_{50}$} & \textbf{AP$^{val}_{75}$} \\	
					\midrule
					\textbf{base (v7-X light)} & 47.0M & 125.5G & 640 & 51.7\% & 70.1\% & 56.0\% \\
					\textbf{width only (1.25 $w$)} & 73.4M & 195.5G & 640 & 52.4\% & 70.9\% & 57.1\%\\
					\textbf{depth only (2.0 $d$)} & 69.3M & 187.6G & 640 & 52.7\% & 70.8\% & 57.3\%\\
					\textbf{compound (v7-X)} & 71.3M & 189.9G & 640 & \textbf{52.9\%} & \textbf{71.1\%} & \textbf{57.5\%}\\
					improvement & - & - & - & \textcolor{green}{+1.2} & \textcolor{green}{+1.0} & \textcolor{green}{+1.5} \\
					\bottomrule
				\end{tabular}
			\end{threeparttable}
		\end{table}
		
		\subsubsection{Proposed planned re-parameterized model}
		
		In order to verify the generality of our proposed planed re-parameterized model, we use it on concatenation-based model and residual-based model respectively for verification. The concatenation-based model and residual-based model we chose for verification are 3-stacked ELAN and CSPDarknet, respectively.
		
\begin{figure}[h]
			\begin{center}
				\includegraphics[width=.9\linewidth]{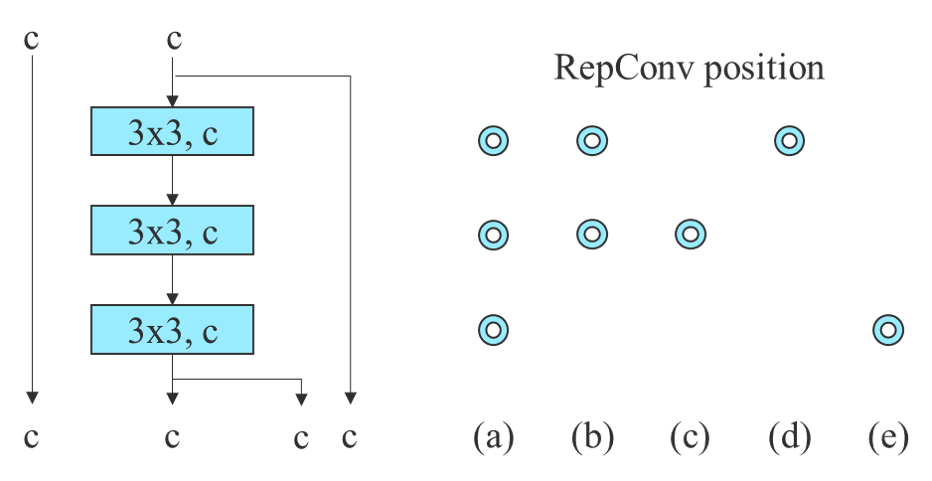}
			\end{center}
			\caption{Planned RepConv 3-stacked ELAN. Blue circles are the position we replace Conv by RepConv.}
			\label{fig:3selan}
		\end{figure}

		In the experiment of concatenation-based model, we replace the $3 \times 3$ convolutional layers in different positions in 3-stacked ELAN with RepConv, and the detailed configuration is shown in Figure \ref{fig:3selan}. From the results shown in Table \ref{table:stack} we see that all higher AP values are present on our proposed planned re-parameterized model.
		
		
		\begin{table}[h]
			\centering
			\begin{threeparttable}[h]
				\footnotesize
				\caption{Ablation study on planned RepConcatenation model.}
				\label{table:stack}
				\setlength\tabcolsep{1.0pt}
				\begin{tabular}{lcccccc}
					\toprule
					\textbf{Model} & \textbf{AP$^{val}$} & \textbf{AP$^{val}_{50}$} & \textbf{AP$^{val}_{75}$} & \textbf{AP$^{val}_{S}$} & \textbf{AP$^{val}_{M}$} & \textbf{AP$^{val}_{L}$}  \\	
					\midrule
					\textbf{base (3-S ELAN)} & 52.26\% & 70.41\% & 56.77\% & 35.81\% & 57.00\% & 67.59\% \\
					\textbf{Figure \ref{fig:3selan} (a)} & 52.18\% & 70.34\% & 56.90\% & 35.71\% & 56.83\% & 67.51\% \\
					\textbf{Figure \ref{fig:3selan} (b)} & 52.30\% & 70.30\% & \textbf{56.92\%} & 35.76\% & 56.95\% & 67.74\% \\
					\textbf{Figure \ref{fig:3selan} (c)} & \textbf{52.33\%} & \textbf{70.56\%} & 56.91\% & \textbf{35.90\%} & \textbf{57.06\%} & 67.50\% \\
					\textbf{Figure \ref{fig:3selan} (d)} & 52.17\% & 70.32\% & 56.82\% & 35.33\% & \textbf{57.06\%} & \textbf{68.09\%} \\
					\textbf{Figure \ref{fig:3selan} (e)} & 52.23\% & 70.20\% & 56.81\% & 35.34\% & 56.97\% & 66.88\% \\
					\bottomrule
				\end{tabular}
			\end{threeparttable}
		\end{table}
		
\begin{figure}[h]
			\begin{center}
				\includegraphics[width=1.\linewidth]{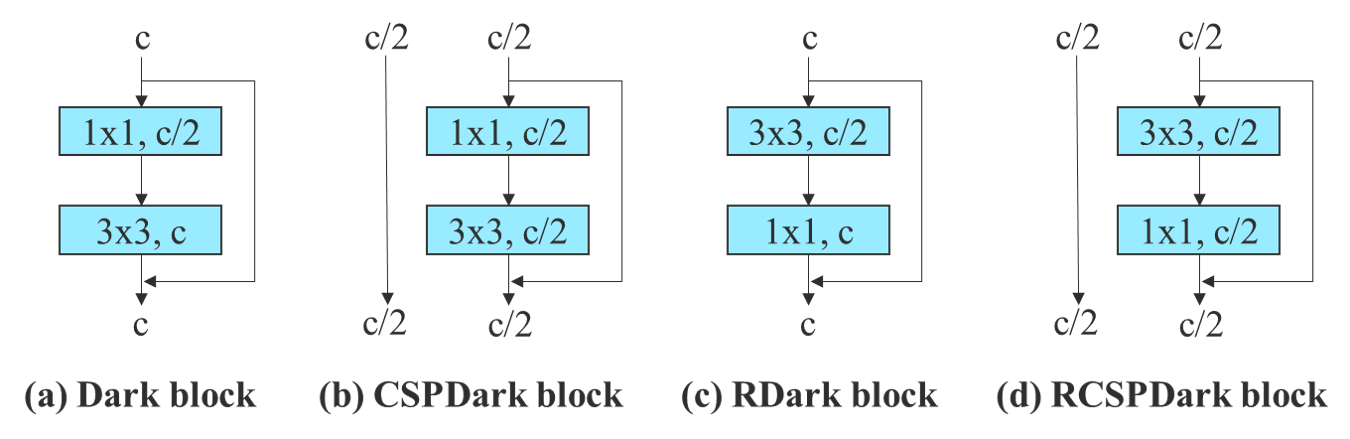}
			\end{center}
			\caption{Reversed CSPDarknet. We reverse the position of $1 \times 1$ and $3 \times 3$ convolutional layer in dark block to fit our planned re-parameterized model design strategy.}
			\label{fig:rcsp}
		\end{figure}

		In the experiment dealing with residual-based model, since the original dark block does not have a $3 \times 3$ convolution block that conforms to our design strategy, we additionally design a reversed dark block for the experiment, whose architecture is shown in Figure \ref{fig:rcsp}. Since the CSPDarknet with dark block and reversed dark block has exactly the same amount of parameters and operations, it is fair to compare. The experiment results illustrated in Table \ref{table:reverse} fully confirm that the proposed planned re-parameterized model is equally effective on residual-based model. We find that the design of RepCSPResNet \cite{xu2022pp} also fit our design pattern.

		\begin{table}[h]
			\centering
			\begin{threeparttable}[h]
				\footnotesize
				\caption{Ablation study on planned RepResidual model.}
				\label{table:reverse}
				\setlength\tabcolsep{.5pt}
				\begin{tabular}{lcccccc}
					\toprule
					\textbf{Model} & \textbf{AP$^{val}$} & \textbf{AP$^{val}_{50}$} & \textbf{AP$^{val}_{75}$} & \textbf{AP$^{val}_{S}$} & \textbf{AP$^{val}_{M}$} & \textbf{AP$^{val}_{L}$}  \\	
					\midrule
					\textbf{base (YOLOR-W6)} & 54.82\% & 72.39\% & 59.95\% & 39.68\% & 59.38\% & \textbf{68.30\%} \\
					\textbf{RepCSP} & 54.67\% & 72.50\% & 59.58\% & 40.22\% & \textbf{59.61\%} & 67.87\% \\
					\textbf{RCSP} & 54.36\% & 71.95\% & 59.54\% & 40.15\% & 59.02\% & 67.44\% \\
					\textbf{RepRCSP} & \textbf{54.85\%} & \textbf{72.51\%} & \textbf{60.08\%} & \textbf{40.53\%} & 59.52\% & 68.06\% \\
					\midrule
					\textbf{base (YOLOR-CSP)} & 50.81\% & 69.47\% & 55.28\% & 33.74\% & \textbf{56.01\%} & 65.38\% \\
					\textbf{RepRCSP} & \textbf{50.91\%} & \textbf{69.54\%} & \textbf{55.55\%} & \textbf{34.44\%} & 55.74\% & \textbf{65.46\%} \\
					\bottomrule
				\end{tabular}
			\end{threeparttable}
		\end{table}
		
		\begin{figure*}[h]
			\begin{center}
				\includegraphics[width=1.0\linewidth]{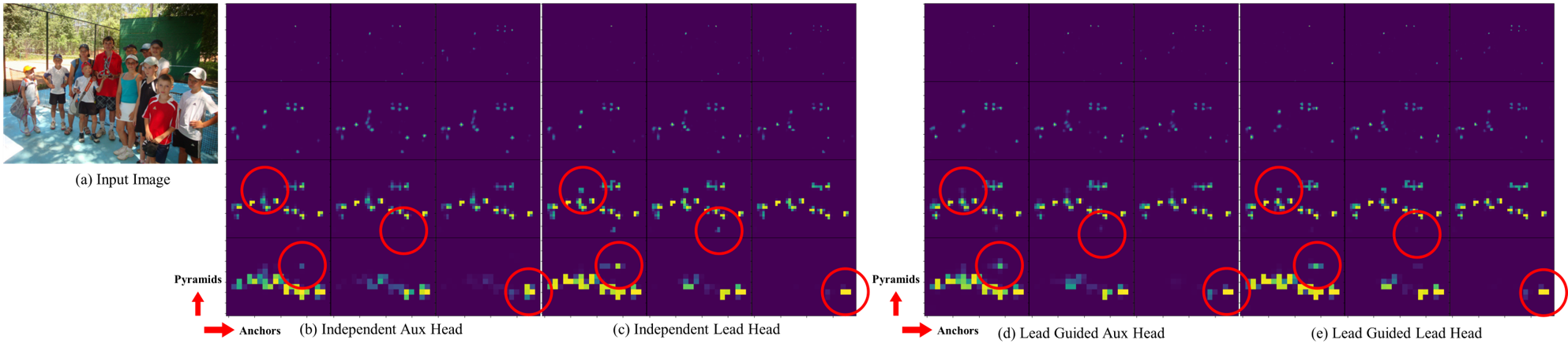}
			\end{center}
			\caption{Objectness map predicted by different methods at auxiliary head and lead head.}
			\label{fig:heat}
			\vspace{-4mm}
		\end{figure*}
		
		\subsubsection{Proposed assistant loss for auxiliary head}
		
		In the assistant loss for auxiliary head experiments, we compare the general independent label assignment for lead head and auxiliary head methods, and we also compare the two proposed lead guided label assignment methods. We show all comparison results in Table \ref{table:aux}. From the results listed in Table \ref{table:aux}, it is clear that any model that increases assistant loss can significantly improve the overall performance. In addition, our proposed lead guided label assignment strategy receives better performance than the general independent label assignment strategy in AP, AP$_{50}$, and AP$_{75}$. As for our proposed coarse for assistant and fine for lead label assignment strategy, it results in best results in all cases. In Figure \ref{fig:heat} we show the objectness map predicted by different methods at auxiliary head and lead head. From Figure \ref{fig:heat} we find that if auxiliary head learns lead guided soft label, it will indeed help lead head to extract the residual information from the consistant targets.
		
		\begin{table}[h]
			\centering
			\begin{threeparttable}[h]
				\footnotesize
				\caption{Ablation study on proposed auxiliary head.}
				\label{table:aux}
				\setlength\tabcolsep{5.0pt}
				\begin{tabular}{lcccc}
					\toprule
					\textbf{Model} & \textbf{Size} & \textbf{AP$^{val}$} & \textbf{AP$^{val}_{50}$} & \textbf{AP$^{val}_{75}$} \\	
					\midrule
					\textbf{base (v7-E6)} & 1280 & 55.6\% & 73.2\% & 60.7\% \\
					\textbf{independent} & 1280 & 55.8\% & 73.4\% & 60.9\%\\
					\textbf{lead guided} & 1280 & 55.9\% & 73.5\% & 61.0\%\\
					\textbf{coarse-to-fine lead guided} & 1280 & \textbf{55.9\%} & \textbf{73.5\%} & \textbf{61.1\%}\\
					improvement & - & \textcolor{green}{+0.3} & \textcolor{green}{+0.3} & \textcolor{green}{+0.4} \\
					\bottomrule
				\end{tabular}
			\end{threeparttable}
		\end{table}
		
		In Table \ref{table:caux} we further analyze the effect of the proposed coarse-to-fine lead guided label assignment method on the decoder of auxiliary head. That is, we compared the results of with/without the introduction of upper bound constraint. Judging from the numbers in the Table, the method of constraining the upper bound of objectness by the distance from the center of the object can achieve better performance.
		
		\begin{table}[h]
			\centering
			\begin{threeparttable}[h]
				\footnotesize
				\caption{Ablation study on constrained auxiliary head.}
				\label{table:caux}
				\setlength\tabcolsep{5.0pt}
				\begin{tabular}{lcccc}
					\toprule
					\textbf{Model} & \textbf{Size} & \textbf{AP$^{val}$} & \textbf{AP$^{val}_{50}$} & \textbf{AP$^{val}_{75}$} \\	
					\midrule
					\textbf{base (v7-E6)} & 1280 & 55.6\% & 73.2\% & 60.7\% \\
					\textbf{aux without constraint} & 1280 & 55.9\% & 73.5\% & 61.0\% \\
					\textbf{aux with constraint} & 1280 & \textbf{55.9\%} & \textbf{73.5\%} & \textbf{61.1\%} \\
					improvement & - & \textcolor{green}{+0.3} & \textcolor{green}{+0.3} & \textcolor{green}{+0.4} \\
					\bottomrule
				\end{tabular}
			\end{threeparttable}
		\end{table}
	
		Since the proposed YOLOv7 uses multiple pyramids to jointly predict object detection results, we can directly connect auxiliary head to the pyramid in the middle layer for training. This type of training can make up for information that may be lost in the next level pyramid prediction. For the above reasons, we designed partial auxiliary head in the proposed E-ELAN architecture. Our approach is to connect auxiliary head after one of the sets of feature map before merging cardinality, and this connection can make the weight of the newly generated set of feature map not directly updated by assistant loss. Our design allows each pyramid of lead head to still get information from objects with different sizes. Table \ref{table:paux} shows the results obtained using two different methods, i.e., coarse-to-fine lead guided and partial coarse-to-fine lead guided methods. Obviously, the partial coarse-to-fine lead guided method has a better auxiliary effect.
		
		\vspace{-2mm}

		\begin{table}[h]
			\centering
			\begin{threeparttable}[h]
				\footnotesize
				\caption{Ablation study on partial auxiliary head.}
				\label{table:paux}
				\setlength\tabcolsep{5.0pt}
				\begin{tabular}{lcccc}
					\toprule
					\textbf{Model} & \textbf{Size} & \textbf{AP$^{val}$} & \textbf{AP$^{val}_{50}$} & \textbf{AP$^{val}_{75}$} \\	
					\midrule
					\textbf{base (v7-E6E)} & 1280 & 56.3\% & 74.0\% & 61.5\% \\
					\textbf{aux} & 1280 & 56.5\% & 74.0\% & 61.6\% \\
					\textbf{partial aux} & 1280 & \textbf{56.8\%} & \textbf{74.4\%} & \textbf{62.1\%} \\
					improvement & - & \textcolor{green}{+0.5} & \textcolor{green}{+0.4} & \textcolor{green}{+0.6} \\
					\bottomrule
				\end{tabular}
			\end{threeparttable}
		\vspace{-4mm}
		\end{table}
	
		\section{Conclusions}
		
		In this paper we propose a new architecture of real-time object detector and the corresponding model scaling method.  Furthermore, we find that the evolving process of object detection methods generates new research topics.  During the research process, we found the replacement problem of re-parameterized module and the allocation problem of dynamic label assignment.  To solve the problem, we propose the trainable bag-of-freebies method to enhance the accuracy of object detection.  Based on the above, we have developed the YOLOv7 series of object detection systems, which receives the state-of-the-art results.

		\section{Acknowledgements}
		
		The authors wish to thank National Center for High-performance Computing (NCHC) for providing computational and storage resources.

		\begin{table*}[t]
			\centering
			\begin{threeparttable}[t]
				\footnotesize
				\caption{More comparison (batch=1, no-TRT, without extra object detection training data)}
				\label{table:more}
				\begin{tabular}{l|c|c|c|c|c|c|c}
					\toprule
					\textbf{Model} & \textbf{\#Param.} & \textbf{FLOPs} & \textbf{Size} & \textbf{FPS$^{V100}$} & \textbf{AP$^{test}$} / \textbf{AP$^{val}$} & \textbf{AP$^{test}_{50}$} & \textbf{AP$^{test}_{75}$} \\
					\midrule
					\textbf{YOLOv7-tiny-SiLU} & 6.2M & 13.8G & 640 & 286 & \textbf{38.7\%} / \textbf{38.7\%} & \textbf{56.7\%} & \textbf{41.7\%} \\
					\textbf{PPYOLOE-S \cite{xu2022pp}} & 7.9M & 17.4G & 640 & 208 &\textbf{ 43.1\%} / \textbf{42.7\%} & \textbf{60.5\%} & \textbf{46.6\%} \\
					\midrule
					\textbf{YOLOv7} & 36.9M & 104.7G & 640 & 161 & \textbf{51.4\%} / \textbf{51.2\%} & \textbf{69.7\%} & \textbf{55.9\%} \\
					\textbf{YOLOv5-N (r6.1) \cite{glenn2022yolov5}} & 1.9M & 4.5G & 640 & 159 & - / 28.0\% & - & - \\
					\textbf{YOLOv5-S (r6.1) \cite{glenn2022yolov5}} & 7.2M & 16.5G & 640 & 156 & - / 37.4\% & - & - \\
					\textbf{PPYOLOE-M \cite{xu2022pp}} & 23.4M & 49.9G & 640 & 123 & 48.9\% / 48.6\% & 66.5\% & 53.0\% \\
					\textbf{YOLOv5-N6 (r6.1) \cite{glenn2022yolov5}} & 3.2M & 18.4G & 1280 & 123 & - / 36.0\% & - & - \\
					\textbf{YOLOv5-S6 (r6.1) \cite{glenn2022yolov5}} & 12.6M & 67.2G & 1280 & 122 & - / 44.8\% & - & -\\
					\textbf{YOLOv5-M (r6.1) \cite{glenn2022yolov5}} & 21.2M & 49.0G & 640 & 122 & - / 45.4\% & - & - \\
					\textbf{YOLOv7-X} & 71.3M & 189.9G & 640 & 114 & \textbf{53.1\%} / \textbf{52.9\%} & \textbf{71.2\%} & \textbf{57.8\%} \\
					\textbf{YOLOR-CSP \cite{wang2021you}} & 52.9M & 120.4G & 640 & 106 & 51.1\% / 50.8\% & 69.6\% & 55.7\% \\
					\textbf{YOLOX-S \cite{ge2021yolox}} & 9.0M & 26.8G & 640 & 102 & 40.5\% / 40.5\% & - & - \\
					\midrule
					\textbf{YOLOv5-L (r6.1) \cite{glenn2022yolov5}} & 46.5M & 109.1G & 640 & 99 & - / 49.0\% & - & - \\
					\textbf{YOLOv5-M6 (r6.1) \cite{glenn2022yolov5}} & 35.7M & 200.0G & 1280 & 90 & - / 51.3\% & - & - \\
					\textbf{YOLOR-CSP-X \cite{wang2021you}} & 96.9M & 226.8G & 640 & 87 & 53.0\% / 52.7\% & \textbf{71.4\%} & \textbf{57.9\%} \\
					\textbf{YOLOv7-W6} & 70.4M & 360.0G & 1280 & 84 & \textbf{54.9\%} / \textbf{54.6\%} & \textbf{72.6\%} & \textbf{60.1\%} \\
					\textbf{YOLOv5-X (r6.1) \cite{glenn2022yolov5}} & 86.7M & 205.7G & 640 & 83 & - / 50.7\% & - & - \\
					\textbf{YOLOX-M \cite{ge2021yolox}} & 25.3M & 73.8G & 640 & 81 & 47.2\% / 46.9\% & - & - \\
					\textbf{PPYOLOE-L \cite{xu2022pp}} & 52.2M & 110.1G & 640 & 78 & 51.4\% / 50.9\% & 68.9\% & 55.6\% \\
					\textbf{YOLOR-P6 \cite{wang2021you}} & 37.2M & 325.6G & 1280 & 76 & 53.9\% / 53.5\% & 71.4\% & 58.9\% \\
					\textbf{YOLOX-L \cite{ge2021yolox}} & 54.2M & 155.6G & 640 & 69 & 50.1\% / 49.7\% & - & - \\
					\textbf{YOLOR-W6 \cite{wang2021you}} & 79.8G & 453.2G & 1280 & 66 & \textbf{55.2\%} / \textbf{54.8\%} & \textbf{72.7\%} & \textbf{60.5\%} \\
					\textbf{YOLOv5-L6 (r6.1) \cite{glenn2022yolov5}} & 76.8M & 445.6G & 1280 & 63 & - / 53.7\% & - & - \\
					\midrule
					\textbf{YOLOX-X \cite{ge2021yolox}} & 99.1M & 281.9G & 640 & 58 & 51.5\% / 51.1\% & - & - \\
					\textbf{YOLOv7-E6} & 97.2M & 515.2G & 1280 & 56 & \textbf{56.0\%} / \textbf{55.9\%} & \textbf{73.5\%} & \textbf{61.2\%} \\
					\textbf{YOLOR-E6 \cite{wang2021you}} & 115.8M & 683.2G & 1280 & 45 & 55.8\% / 55.7\% & 73.4\% & 61.1\% \\
					\textbf{PPYOLOE-X \cite{xu2022pp}} & 98.4M & 206.6G & 640 & 45 & 52.2\% / 51.9\% & 69.9\% & 56.5\% \\
					\textbf{YOLOv7-D6} & 154.7M & 806.8G & 1280 & 44 & \textbf{56.6\%} / \textbf{56.3\%} & \textbf{74.0\%} & \textbf{61.8\%} \\
					\textbf{YOLOv5-X6 (r6.1) \cite{glenn2022yolov5}} & 140.7M & 839.2G & 1280 & 38 & - / 55.0\% & - & - \\
					\textbf{YOLOv7-E6E} & 151.7M & 843.2G & 1280 & 36 & \textbf{56.8\%} / \textbf{56.8\%} & \textbf{74.4\%} & \textbf{62.1\%} \\
					\textbf{YOLOR-D6 \cite{wang2021you}} & 151.7M & 935.6G & 1280 & 34 & 56.5\% / 56.1\% & \textbf{74.1\%} & \textbf{61.9\%} \\
					\midrule
					\textbf{F-RCNN-R101-FPN+ \cite{carion2020end}} & 60.0M & 246.0G & 1333 &  20 & - / 44.0\% & - & - \\
					\textbf{Deformable DETR \cite{zhu2021deformable}} & 40.0M & 173.0G & - &  19 & - / 46.2\% & - & - \\
					\textbf{Swin-B (C-M-RCNN) \cite{liu2021swin}} & 145.0M & 982.0G & 1333 &  11.6 & - / 51.9\% & - & - \\
					\textbf{DETR DC5-R101 \cite{carion2020end}} & 60.0M & 253.0G & 1333 &  10 & - / 44.9\% & - & - \\
					\textbf{EfficientDet-D7x \cite{tan2020efficientdet}} & 77.0M & 410.0G & 1536 &  6.5 & 55.1\% / 54.4\% &  72.4\% & 58.4\% \\
					\textbf{Dual-Swin-T (C-M-RCNN) \cite{liang2021cbnetv2}} & 113.8M & 836.0G & 1333 &  6.5 & - / 53.6\% & - & - \\
					\textbf{ViT-Adapter-B \cite{chen2022vision}} & 122.0M & 997.0G & - & 4.4 & - / 50.8\% & - & - \\
					\textbf{Dual-Swin-B (HTC) \cite{liang2021cbnetv2}} & 235.0M & - & 1600 & 2.5 & \textbf{58.7\%} / \textbf{58.4\%} & - & - \\
					\textbf{Dual-Swin-L (HTC) \cite{liang2021cbnetv2}} & 453.0M & - & 1600 & 1.5 & \textbf{59.4\%} / \textbf{59.1\%} & - & - \\
					\bottomrule
					\toprule
					\textbf{Model} & \textbf{\#Param.} & \textbf{FLOPs} & \textbf{Size} & \textbf{FPS$^{A100}$} & \textbf{AP$^{test}$} / \textbf{AP$^{val}$} & \textbf{AP$^{test}_{50}$} & \textbf{AP$^{test}_{75}$} \\
					\midrule
					\textbf{DN-Deformable-DETR \cite{li2022dn}} & 48.0M & 265.0G & 1333 &  23.0 & - / 48.6\% & - & - \\
					\textbf{ConvNeXt-B (C-M-RCNN) \cite{liu2022convnext}} & - & 964.0G & 1280 &  11.5 & - / 54.0\% & 73.1\% & 58.8\% \\
					\textbf{Swin-B (C-M-RCNN) \cite{liu2021swin}} & - & 982.0G & 1280 &  10.7 & - / 53.0\% & 71.8\% & 57.5\% \\
					\textbf{DINO-5scale (R50) \cite{zhang2022dino}} & 47.0M & 860.0G & 1333 &  10.0 & - / 51.0\% & - & - \\
					\textbf{ConvNeXt-L (C-M-RCNN) \cite{liu2022convnext}} & - & 1354.0G & 1280 &  10.0 & - / 54.8\% & 73.8\% & 59.8\% \\
					\textbf{Swin-L (C-M-RCNN) \cite{liu2021swin}} & - & 1382.0G & 1280 &  9.2 & - / 53.9\% & 72.4\% & 58.8\% \\
					\textbf{ConvNeXt-XL (C-M-RCNN) \cite{liu2022convnext}} & - & 1898.0G & 1280 &  8.6 & - / 55.2\% & 74.2\% & 59.9\% \\
					\bottomrule
				\end{tabular}
			\end{threeparttable}
		\vspace{-3mm}
		\end{table*}
	
		\section{More comparison}
		
		\vspace{-3mm}
		
		YOLOv7 surpasses all known object detectors in both speed and accuracy in the range from 5 FPS to 160 FPS and has the highest accuracy 56.8\% AP test-dev / 56.8\% AP min-val among all known real-time object detectors with 30 FPS or higher on GPU V100. YOLOv7-E6 object detector (56 FPS V100, 55.9\% AP) outperforms both transformer-based detector SWIN-L Cascade-Mask R-CNN (9.2 FPS A100, 53.9\% AP) by 509\% in speed and 2\% in accuracy, and convolutional-based detector ConvNeXt-XL Cascade-Mask R-CNN (8.6 FPS A100, 55.2\% AP) by 551\% in speed and 0.7\% AP in accuracy, as well as YOLOv7 outperforms: YOLOR, YOLOX, Scaled-YOLOv4, YOLOv5, DETR, Deformable DETR, DINO-5scale-R50, ViT-Adapter-B and many other object detectors in speed and accuracy. More over, we train YOLOv7 only on MS COCO dataset from scratch without using any other datasets or pre-trained weights.

		\begin{figure*}[h]
			\begin{center}
				\includegraphics[width=.85\linewidth]{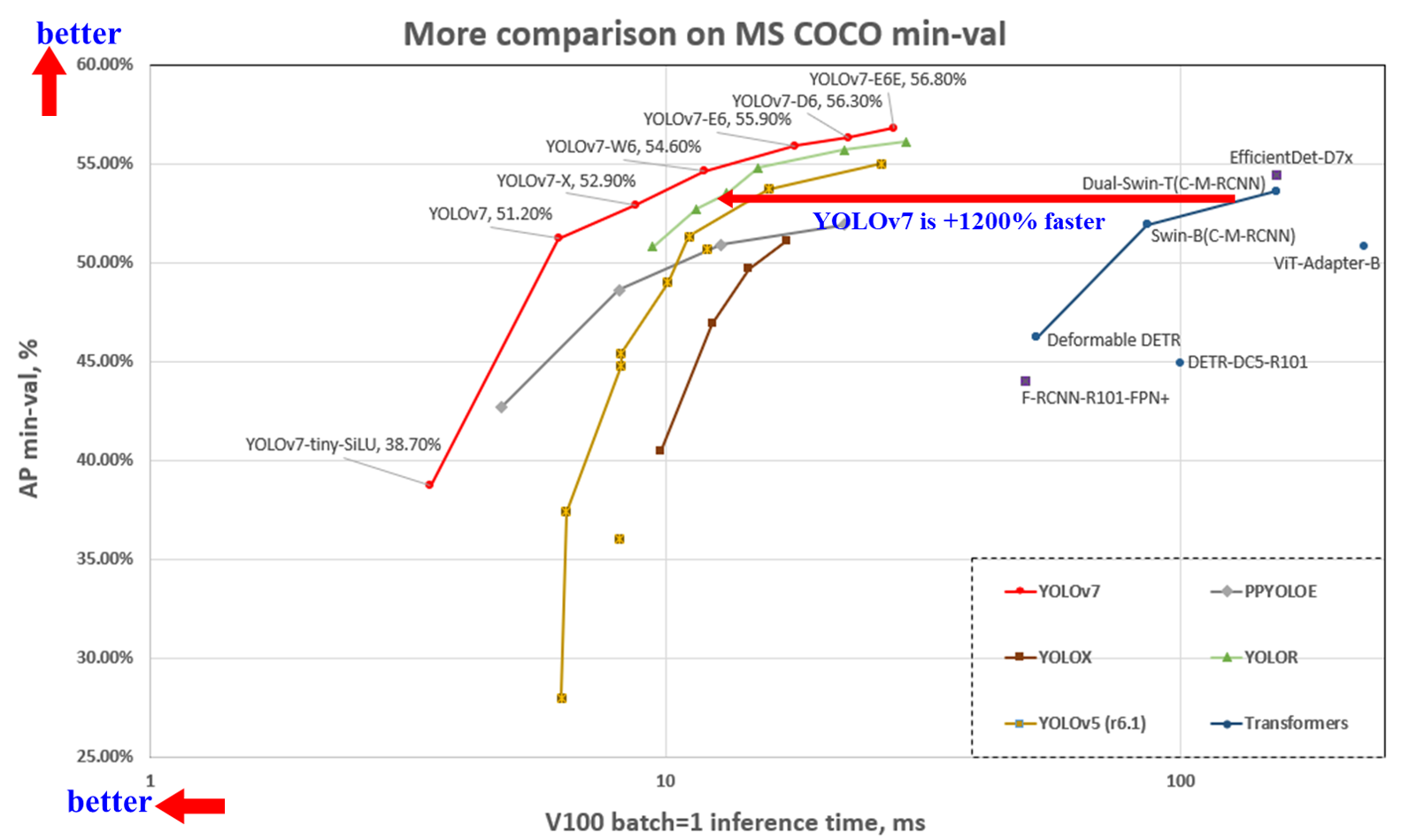}
			\end{center}
			\caption{Comparison with other object detectors.}
			\label{fig:more}
			\vspace{-4mm}
		\end{figure*}

		\begin{figure*}[h]
			\begin{center}
				\includegraphics[width=.85\linewidth]{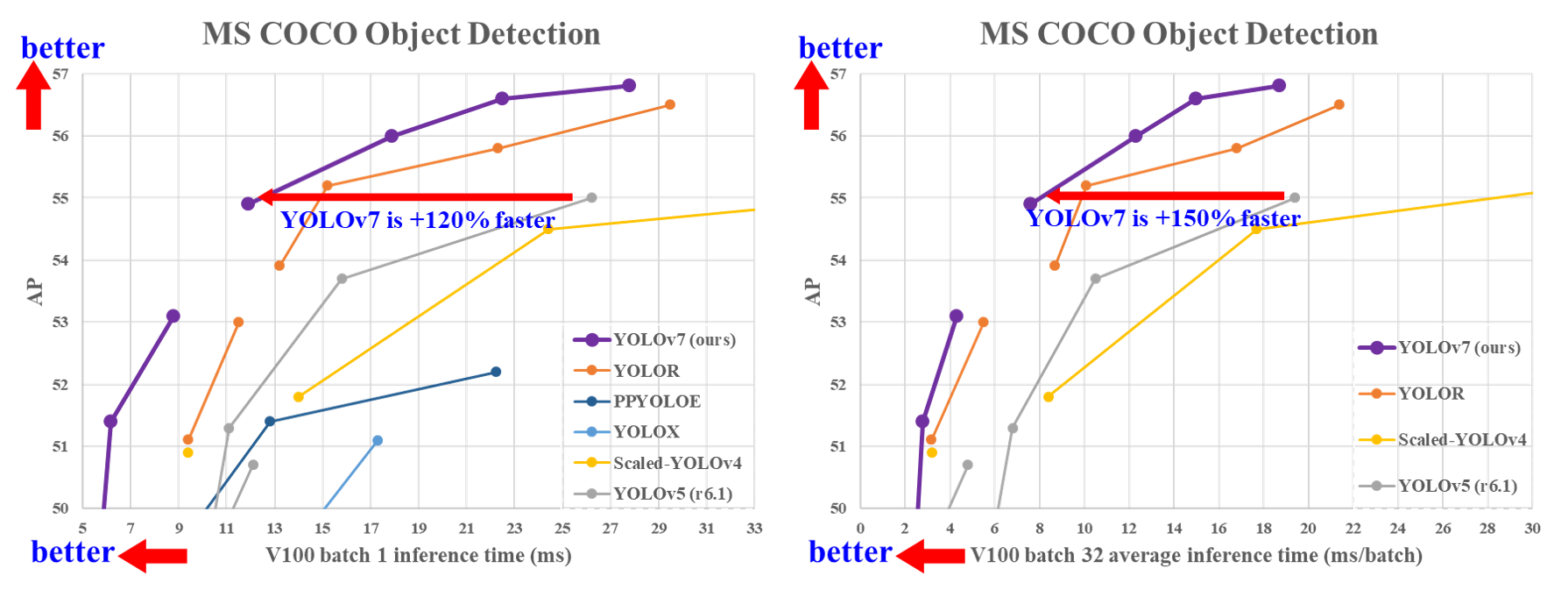}
			\end{center}
			\caption{Comparison with other real-time object detectors.}
			\label{fig:sota32}
			\vspace{-4mm}
		\end{figure*}
	
		\begin{table}[h]
		\centering
		\begin{threeparttable}[h]
			\footnotesize
			\caption{Comparison of different setting.}
			\label{table:set}
			\begin{tabular}{lccc}
				\toprule
				\textbf{Model} & \textbf{Presicion} & \textbf{IoU threshold} & \textbf{AP$^{val}$} \\	
				\midrule
				\textbf{YOLOv7-X} & FP16 (default) & 0.65 (default) & \textbf{52.9\%}  \\
				\textbf{YOLOv7-X} & FP32 & 0.65 & \textbf{53.0\%}  \\
				\textbf{YOLOv7-X} & FP16 & 0.70 & \textbf{53.0\%}  \\
				\textbf{YOLOv7-X} & FP32 & 0.70 & \textbf{53.1\%}  \\
				improvement & - & - & \textcolor{green}{+0.2\%}  \\
				\bottomrule
			\end{tabular}
			\begin{tablenotes}[flushleft]
			\footnotesize
			\item[*] Similar to meituan/YOLOv6 and PPYOLOE, our model could get higher AP when set higher IoU threshold.
			\end{tablenotes}
		\end{threeparttable}
		\vspace{-5mm}
		\end{table}
	
		The maximum accuracy of the YOLOv7-E6E (56.8\% AP) real-time model is +13.7\% AP higher than the current most accurate meituan/YOLOv6-s model (43.1\% AP) on COCO dataset.	Our YOLOv7-tiny (35.2\% AP, 0.4 ms) model is +25\% faster and +0.2\% AP higher than meituan/YOLOv6-n (35.0\% AP, 0.5 ms) under identical conditions on COCO dataset and V100 GPU with batch=32.

		\begin{figure}[h]
			\begin{center}
				\includegraphics[width=.9\linewidth]{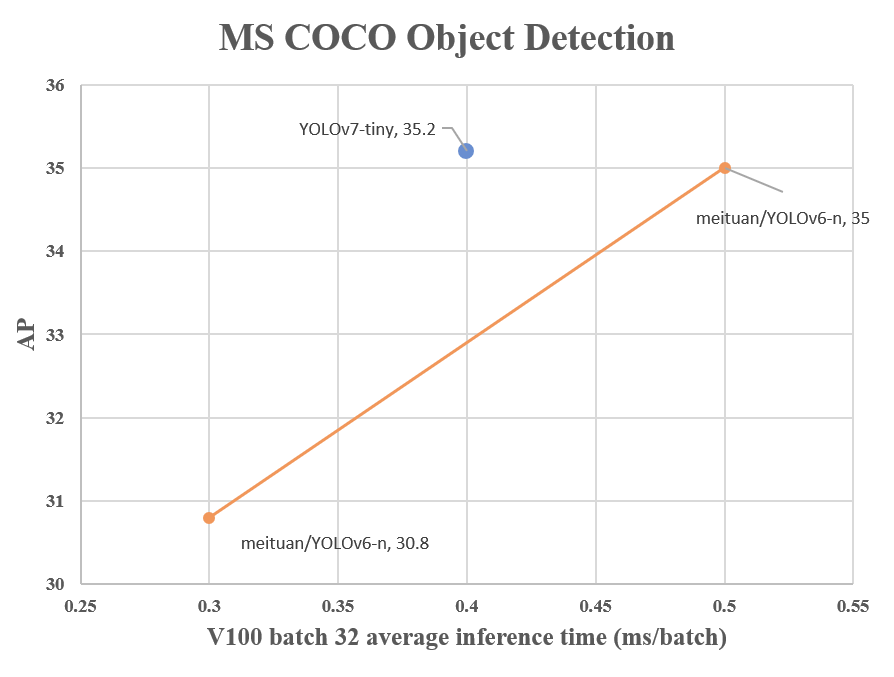}
			\end{center}
			\caption{Comparison with other real-time object detectors.}
			\label{fig:mt}
		\end{figure}

		%
		%
		
		\clearpage
		\clearpage
		\clearpage
		
				
		\newpage
		
		\appendix
		
		\setcounter{table}{0}
		\renewcommand{\thetable}{A\arabic{table}}
		
		\setcounter{figure}{0}
		\renewcommand{\thefigure}{A\arabic{figure}}
		
		\clearpage
		




{\small
			\begin{thebibliography}{100}\itemsep=-1pt
				
				\bibitem{wang2022designing}
				anonymous.
				\newblock Designing network design strategies.
				\newblock {\em anonymous submission}, 2022.
				
				\bibitem{bello2021revisiting}
				Irwan Bello, William Fedus, Xianzhi Du, Ekin~Dogus Cubuk, Aravind Srinivas,
				Tsung-Yi Lin, Jonathon Shlens, and Barret Zoph.
				\newblock Revisiting {ResNets}: Improved training and scaling strategies.
				\newblock {\em Advances in Neural Information Processing Systems (NeurIPS)},
				34, 2021.
				
				\bibitem{bochkovskiy2020yolov4}
				Alexey Bochkovskiy, Chien-Yao Wang, and Hong-Yuan~Mark Liao.
				\newblock {YOLOv4}: Optimal speed and accuracy of object detection.
				\newblock {\em arXiv preprint arXiv:2004.10934}, 2020.
				
				\bibitem{cao2020ensemble}
				Yue Cao, Thomas~Andrew Geddes, Jean Yee~Hwa Yang, and Pengyi Yang.
				\newblock Ensemble deep learning in bioinformatics.
				\newblock {\em Nature Machine Intelligence}, 2(9):500--508, 2020.
				
				\bibitem{carion2020end}
				Nicolas Carion, Francisco Massa, Gabriel Synnaeve, Nicolas Usunier, Alexander
				Kirillov, and Sergey Zagoruyko.
				\newblock End-to-end object detection with transformers.
				\newblock In {\em Proceedings of the European Conference on Computer Vision
					(ECCV)}, pages 213--229, 2020.
				
				\bibitem{chen2020ap}
				Kean Chen, Weiyao Lin, Jianguo Li, John See, Ji Wang, and Junni Zou.
				\newblock {AP}-loss for accurate one-stage object detection.
				\newblock {\em IEEE Transactions on Pattern Analysis and Machine Intelligence
					(TPAMI)}, 43(11):3782--3798, 2020.
				
				\bibitem{chen2022vision}
				Zhe Chen, Yuchen Duan, Wenhai Wang, Junjun He, Tong Lu, Jifeng Dai, and Yu
				Qiao.
				\newblock Vision transformer adapter for dense predictions.
				\newblock {\em arXiv preprint arXiv:2205.08534}, 2022.
				
				\bibitem{choi2019gaussian}
				Jiwoong Choi, Dayoung Chun, Hyun Kim, and Hyuk-Jae Lee.
				\newblock {Gaussian YOLOv3}: An accurate and fast object detector using
				localization uncertainty for autonomous driving.
				\newblock In {\em Proceedings of the IEEE/CVF International Conference on
					Computer Vision (ICCV)}, pages 502--511, 2019.
				
				\bibitem{dai2021dynamic}
				Xiyang Dai, Yinpeng Chen, Bin Xiao, Dongdong Chen, Mengchen Liu, Lu Yuan, and
				Lei Zhang.
				\newblock Dynamic head: Unifying object detection heads with attentions.
				\newblock In {\em Proceedings of the IEEE/CVF Conference on Computer Vision and
					Pattern Recognition (CVPR)}, pages 7373--7382, 2021.
				
				\bibitem{ding2022re}
				Xiaohan Ding, Honghao Chen, Xiangyu Zhang, Kaiqi Huang, Jungong Han, and
				Guiguang Ding.
				\newblock Re-parameterizing your optimizers rather than architectures.
				\newblock {\em arXiv preprint arXiv:2205.15242}, 2022.
				
				\bibitem{ding2019acnet}
				Xiaohan Ding, Yuchen Guo, Guiguang Ding, and Jungong Han.
				\newblock {ACNet}: Strengthening the kernel skeletons for powerful {CNN} via
				asymmetric convolution blocks.
				\newblock In {\em Proceedings of the IEEE/CVF International Conference on
					Computer Vision (ICCV)}, pages 1911--1920, 2019.
				
				\bibitem{ding2021diverse}
				Xiaohan Ding, Xiangyu Zhang, Jungong Han, and Guiguang Ding.
				\newblock Diverse branch block: Building a convolution as an inception-like
				unit.
				\newblock In {\em Proceedings of the IEEE/CVF Conference on Computer Vision and
					Pattern Recognition (CVPR)}, pages 10886--10895, 2021.
				
				\bibitem{ding2021repvgg}
				Xiaohan Ding, Xiangyu Zhang, Ningning Ma, Jungong Han, Guiguang Ding, and Jian
				Sun.
				\newblock {RepVGG}: Making {VGG}-style convnets great again.
				\newblock In {\em Proceedings of the IEEE/CVF Conference on Computer Vision and
					Pattern Recognition (CVPR)}, pages 13733--13742, 2021.
				
				\bibitem{ding2022scaling}
				Xiaohan Ding, Xiangyu Zhang, Yizhuang Zhou, Jungong Han, Guiguang Ding, and
				Jian Sun.
				\newblock Scaling up your kernels to 31x31: Revisiting large kernel design in
				{CNNs}.
				\newblock In {\em Proceedings of the IEEE/CVF Conference on Computer Vision and
					Pattern Recognition (CVPR)}, 2022.
				
				\bibitem{dollar2021fast}
				Piotr Doll{\'a}r, Mannat Singh, and Ross Girshick.
				\newblock Fast and accurate model scaling.
				\newblock In {\em Proceedings of the IEEE/CVF Conference on Computer Vision and
					Pattern Recognition (CVPR)}, pages 924--932, 2021.
				
				\bibitem{du2021simple}
				Xianzhi Du, Barret Zoph, Wei-Chih Hung, and Tsung-Yi Lin.
				\newblock Simple training strategies and model scaling for object detection.
				\newblock {\em arXiv preprint arXiv:2107.00057}, 2021.
				
				\bibitem{feng2021tood}
				Chengjian Feng, Yujie Zhong, Yu Gao, Matthew~R Scott, and Weilin Huang.
				\newblock {TOOD}: Task-aligned one-stage object detection.
				\newblock In {\em Proceedings of the IEEE/CVF International Conference on
					Computer Vision (ICCV)}, pages 3490--3499, 2021.
				
				\bibitem{feng2020deep}
				Di Feng, Christian Haase-Sch{\"u}tz, Lars Rosenbaum, Heinz Hertlein, Claudius
				Glaeser, Fabian Timm, Werner Wiesbeck, and Klaus Dietmayer.
				\newblock Deep multi-modal object detection and semantic segmentation for
				autonomous driving: Datasets, methods, and challenges.
				\newblock {\em IEEE Transactions on Intelligent Transportation Systems},
				22(3):1341--1360, 2020.
				
				\bibitem{garipov2018loss}
				Timur Garipov, Pavel Izmailov, Dmitrii Podoprikhin, Dmitry~P Vetrov, and
				Andrew~G Wilson.
				\newblock Loss surfaces, mode connectivity, and fast ensembling of {DNNs}.
				\newblock {\em Advances in Neural Information Processing Systems (NeurIPS)},
				31, 2018.
				
				\bibitem{ge2021ota}
				Zheng Ge, Songtao Liu, Zeming Li, Osamu Yoshie, and Jian Sun.
				\newblock {OTA}: Optimal transport assignment for object detection.
				\newblock In {\em Proceedings of the IEEE/CVF Conference on Computer Vision and
					Pattern Recognition (CVPR)}, pages 303--312, 2021.
				
				\bibitem{ge2021yolox}
				Zheng Ge, Songtao Liu, Feng Wang, Zeming Li, and Jian Sun.
				\newblock {YOLOX}: Exceeding {YOLO} series in 2021.
				\newblock {\em arXiv preprint arXiv:2107.08430}, 2021.
				
				\bibitem{ghiasi2019fpn}
				Golnaz Ghiasi, Tsung-Yi Lin, and Quoc~V Le.
				\newblock {NAS-FPN}: Learning scalable feature pyramid architecture for object
				detection.
				\newblock In {\em Proceedings of the IEEE/CVF Conference on Computer Vision and
					Pattern Recognition (CVPR)}, pages 7036--7045, 2019.
				
				\bibitem{glenn2022yolov5}
				Jocher Glenn.
				\newblock {YOLOv5} release v6.1.
				\newblock \url{https://github.com/ultralytics/yolov5/releases/tag/v6.1}, 2022.
				
				\bibitem{guo2020expandnets}
				Shuxuan Guo, Jose~M Alvarez, and Mathieu Salzmann.
				\newblock {ExpandNets}: Linear over-parameterization to train compact
				convolutional networks.
				\newblock {\em Advances in Neural Information Processing Systems (NeurIPS)},
				33:1298--1310, 2020.
				
				\bibitem{han2020ghostnet}
				Kai Han, Yunhe Wang, Qi Tian, Jianyuan Guo, Chunjing Xu, and Chang Xu.
				\newblock {GhostNet}: More features from cheap operations.
				\newblock In {\em Proceedings of the IEEE/CVF Conference on Computer Vision and
					Pattern Recognition (CVPR)}, pages 1580--1589, 2020.
				
				\bibitem{he2016deep}
				Kaiming He, Xiangyu Zhang, Shaoqing Ren, and Jian Sun.
				\newblock Deep residual learning for image recognition.
				\newblock In {\em Proceedings of the IEEE/CVF Conference on Computer Vision and
					Pattern Recognition (CVPR)}, pages 770--778, 2016.
				
				\bibitem{howard2019searching}
				Andrew Howard, Mark Sandler, Grace Chu, Liang-Chieh Chen, Bo Chen, Mingxing
				Tan, Weijun Wang, Yukun Zhu, Ruoming Pang, Vijay Vasudevan, et~al.
				\newblock Searching for {MobileNetV3}.
				\newblock In {\em Proceedings of the IEEE/CVF Conference on Computer Vision and
					Pattern Recognition (CVPR)}, pages 1314--1324, 2019.
				
				\bibitem{howard2017mobilenets}
				Andrew~G Howard, Menglong Zhu, Bo Chen, Dmitry Kalenichenko, Weijun Wang,
				Tobias Weyand, Marco Andreetto, and Hartwig Adam.
				\newblock {MobileNets}: Efficient convolutional neural networks for mobile
				vision applications.
				\newblock {\em arXiv preprint arXiv:1704.04861}, 2017.
				
				\bibitem{hu2022online}
				Mu Hu, Junyi Feng, Jiashen Hua, Baisheng Lai, Jianqiang Huang, Xiaojin Gong,
				and Xiansheng Hua.
				\newblock Online convolutional re-parameterization.
				\newblock In {\em Proceedings of the IEEE/CVF Conference on Computer Vision and
					Pattern Recognition (CVPR)}, 2022.
				
				\bibitem{hu2021a2}
				Miao Hu, Yali Li, Lu Fang, and Shengjin Wang.
				\newblock {A$^{2}$-FPN}: Attention aggregation based feature pyramid network
				for instance segmentation.
				\newblock In {\em Proceedings of the IEEE/CVF Conference on Computer Vision and
					Pattern Recognition (CVPR)}, pages 15343--15352, 2021.
				
				\bibitem{huang2017snapshot}
				Gao Huang, Yixuan Li, Geoff Pleiss, Zhuang Liu, John~E Hopcroft, and Kilian~Q
				Weinberger.
				\newblock Snapshot ensembles: Train 1, get m for free.
				\newblock {\em International Conference on Learning Representations (ICLR)},
				2017.
				
				\bibitem{huang2017densely}
				Gao Huang, Zhuang Liu, Laurens Van Der~Maaten, and Kilian~Q Weinberger.
				\newblock Densely connected convolutional networks.
				\newblock In {\em Proceedings of the IEEE/CVF Conference on Computer Vision and
					Pattern Recognition (CVPR)}, pages 4700--4708, 2017.
				
				\bibitem{izmailov2018averaging}
				Pavel Izmailov, Dmitrii Podoprikhin, Timur Garipov, Dmitry Vetrov, and
				Andrew~Gordon Wilson.
				\newblock Averaging weights leads to wider optima and better generalization.
				\newblock In {\em Conference on Uncertainty in Artificial Intelligence (UAI)},
				2018.
				
				\bibitem{jaeger2020retina}
				Paul~F Jaeger, Simon~AA Kohl, Sebastian Bickelhaupt, Fabian Isensee,
				Tristan~Anselm Kuder, Heinz-Peter Schlemmer, and Klaus~H Maier-Hein.
				\newblock {Retina U-Net}: Embarrassingly simple exploitation of segmentation
				supervision for medical object detection.
				\newblock In {\em Machine Learning for Health Workshop}, pages 171--183, 2020.
				
				\bibitem{karaoguz2019object}
				Hakan Karaoguz and Patric Jensfelt.
				\newblock Object detection approach for robot grasp detection.
				\newblock In {\em IEEE International Conference on Robotics and Automation
					(ICRA)}, pages 4953--4959, 2019.
				
				\bibitem{kim2020probabilistic}
				Kang Kim and Hee~Seok Lee.
				\newblock Probabilistic anchor assignment with iou prediction for object
				detection.
				\newblock In {\em Proceedings of the European conference on computer vision
					(ECCV)}, pages 355--371, 2020.
				
				\bibitem{kirillov2019panoptic}
				Alexander Kirillov, Ross Girshick, Kaiming He, and Piotr Doll{\'a}r.
				\newblock Panoptic feature pyramid networks.
				\newblock In {\em Proceedings of the IEEE/CVF Conference on Computer Vision and
					Pattern Recognition (CVPR)}, pages 6399--6408, 2019.
				
				\bibitem{lee2015deeply}
				Chen-Yu Lee, Saining Xie, Patrick Gallagher, Zhengyou Zhang, and Zhuowen Tu.
				\newblock Deeply-supervised nets.
				\newblock In {\em Artificial Intelligence and Statistics}, pages 562--570,
				2015.
				
				\bibitem{lee2019energy}
				Youngwan Lee, Joong-won Hwang, Sangrok Lee, Yuseok Bae, and Jongyoul Park.
				\newblock An energy and {GPU}-computation efficient backbone network for
				real-time object detection.
				\newblock In {\em Proceedings of the IEEE/CVF Conference on Computer Vision and
					Pattern Recognition Workshops (CVPRW)}, pages 0--0, 2019.
				
				\bibitem{li2019gs3d}
				Buyu Li, Wanli Ouyang, Lu Sheng, Xingyu Zeng, and Xiaogang Wang.
				\newblock {GS3D}: An efficient 3d object detection framework for autonomous
				driving.
				\newblock In {\em Proceedings of the IEEE/CVF Conference on Computer Vision and
					Pattern Recognition (CVPR)}, pages 1019--1028, 2019.
				
				\bibitem{li2022dn}
				Feng Li, Hao Zhang, Shilong Liu, Jian Guo, Lionel~M Ni, and Lei Zhang.
				\newblock {DN-DETR}: Accelerate detr training by introducing query denoising.
				\newblock In {\em Proceedings of the IEEE/CVF Conference on Computer Vision and
					Pattern Recognition (CVPR)}, pages 13619--13627, 2022.
				
				\bibitem{li2022dual}
				Shuai Li, Chenhang He, Ruihuang Li, and Lei Zhang.
				\newblock A dual weighting label assignment scheme for object detection.
				\newblock In {\em Proceedings of the IEEE/CVF Conference on Computer Vision and
					Pattern Recognition (CVPR)}, pages 9387--9396, 2022.
				
				\bibitem{li2021generalized}
				Xiang Li, Wenhai Wang, Xiaolin Hu, Jun Li, Jinhui Tang, and Jian Yang.
				\newblock Generalized focal loss v2: Learning reliable localization quality
				estimation for dense object detection.
				\newblock In {\em Proceedings of the IEEE/CVF Conference on Computer Vision and
					Pattern Recognition (CVPR)}, pages 11632--11641, 2021.
				
				\bibitem{li2020generalized}
				Xiang Li, Wenhai Wang, Lijun Wu, Shuo Chen, Xiaolin Hu, Jun Li, Jinhui Tang,
				and Jian Yang.
				\newblock Generalized focal loss: Learning qualified and distributed bounding
				boxes for dense object detection.
				\newblock {\em Advances in Neural Information Processing Systems (NeurIPS)},
				33:21002--21012, 2020.
				
				\bibitem{li2022exploring}
				Yanghao Li, Hanzi Mao, Ross Girshick, and Kaiming He.
				\newblock Exploring plain vision transformer backbones for object detection.
				\newblock {\em arXiv preprint arXiv:2203.16527}, 2022.
				
				\bibitem{li2019clu}
				Zhuoling Li, Minghui Dong, Shiping Wen, Xiang Hu, Pan Zhou, and Zhigang Zeng.
				\newblock {CLU-CNNs}: Object detection for medical images.
				\newblock {\em Neurocomputing}, 350:53--59, 2019.
				
				\bibitem{liang2021cbnetv2}
				Tingting Liang, Xiaojie Chu, Yudong Liu, Yongtao Wang, Zhi Tang, Wei Chu,
				Jingdong Chen, and Haibin Ling.
				\newblock {CBNetV2}: A composite backbone network architecture for object
				detection.
				\newblock {\em arXiv preprint arXiv:2107.00420}, 2021.
				
				\bibitem{lin2021memory}
				Ji Lin, Wei-Ming Chen, Han Cai, Chuang Gan, and Song Han.
				\newblock Memory-efficient patch-based inference for tiny deep learning.
				\newblock {\em Advances in Neural Information Processing Systems (NeurIPS)},
				34:2346--2358, 2021.
				
				\bibitem{lin2020mcunet}
				Ji Lin, Wei-Ming Chen, Yujun Lin, Chuang Gan, Song Han, et~al.
				\newblock {MCUNet}: Tiny deep learning on {IoT} devices.
				\newblock {\em Advances in Neural Information Processing Systems (NeurIPS)},
				33:11711--11722, 2020.
				
				\bibitem{liu2021yolostereo3d}
				Yuxuan Liu, Lujia Wang, and Ming Liu.
				\newblock {YOLOStereo3D}: A step back to {2D} for efficient stereo {3D}
				detection.
				\newblock In {\em IEEE International Conference on Robotics and Automation
					(ICRA)}, pages 13018--13024, 2021.
				
				\bibitem{liu2022swin}
				Ze Liu, Han Hu, Yutong Lin, Zhuliang Yao, Zhenda Xie, Yixuan Wei, Jia Ning, Yue
				Cao, Zheng Zhang, Li Dong, et~al.
				\newblock Swin transformer v2: Scaling up capacity and resolution.
				\newblock In {\em Proceedings of the IEEE/CVF Conference on Computer Vision and
					Pattern Recognition (CVPR)}, 2022.
				
				\bibitem{liu2021swin}
				Ze Liu, Yutong Lin, Yue Cao, Han Hu, Yixuan Wei, Zheng Zhang, Stephen Lin, and
				Baining Guo.
				\newblock Swin transformer: Hierarchical vision transformer using shifted
				windows.
				\newblock In {\em Proceedings of the IEEE/CVF International Conference on
					Computer Vision (ICCV)}, pages 10012--10022, 2021.
				
				\bibitem{liu2022convnext}
				Zhuang Liu, Hanzi Mao, Chao-Yuan Wu, Christoph Feichtenhofer, Trevor Darrell,
				and Saining Xie.
				\newblock A {ConvNet} for the 2020s.
				\newblock In {\em Proceedings of the IEEE/CVF Conference on Computer Vision and
					Pattern Recognition (CVPR)}, pages 11976--11986, 2022.
				
				\bibitem{lyu2021nano}
				Rangi Lyu.
				\newblock {NanoDet-Plus}.
				\newblock
				\url{https://github.com/RangiLyu/nanodet/releases/tag/v1.0.0-alpha-1}, 2021.
				
				\bibitem{ma2018shufflenetv2}
				Ningning Ma, Xiangyu Zhang, Hai-Tao Zheng, and Jian Sun.
				\newblock {ShuffleNet V2}: Practical guidelines for efficient {CNN}
				architecture design.
				\newblock In {\em Proceedings of the European Conference on Computer Vision
					(ECCV)}, pages 116--131, 2018.
				
				\bibitem{oksuz2020ranking}
				Kemal Oksuz, Baris~Can Cam, Emre Akbas, and Sinan Kalkan.
				\newblock A ranking-based, balanced loss function unifying classification and
				localisation in object detection.
				\newblock {\em Advances in Neural Information Processing Systems (NeurIPS)},
				33:15534--15545, 2020.
				
				\bibitem{oksuz2021rank}
				Kemal Oksuz, Baris~Can Cam, Emre Akbas, and Sinan Kalkan.
				\newblock Rank \& sort loss for object detection and instance segmentation.
				\newblock In {\em Proceedings of the IEEE/CVF International Conference on
					Computer Vision (ICCV)}, pages 3009--3018, 2021.
				
				\bibitem{paul2021object}
				Shuvo~Kumar Paul, Muhammed~Tawfiq Chowdhury, Mircea Nicolescu, Monica
				Nicolescu, and David Feil-Seifer.
				\newblock Object detection and pose estimation from rgb and depth data for
				real-time, adaptive robotic grasping.
				\newblock In {\em Advances in Computer Vision and Computational Biology}, pages
				121--142. 2021.
				
				\bibitem{qiao2021detectors}
				Siyuan Qiao, Liang-Chieh Chen, and Alan Yuille.
				\newblock {DetectoRS}: Detecting objects with recursive feature pyramid and
				switchable atrous convolution.
				\newblock In {\em Proceedings of the IEEE/CVF Conference on Computer Vision and
					Pattern Recognition (CVPR)}, pages 10213--10224, 2021.
				
				\bibitem{radosavovic2020designing}
				Ilija Radosavovic, Raj~Prateek Kosaraju, Ross Girshick, Kaiming He, and Piotr
				Doll{\'a}r.
				\newblock Designing network design spaces.
				\newblock In {\em Proceedings of the IEEE/CVF Conference on Computer Vision and
					Pattern Recognition (CVPR)}, pages 10428--10436, 2020.
				
				\bibitem{redmon2016you}
				Joseph Redmon, Santosh Divvala, Ross Girshick, and Ali Farhadi.
				\newblock You only look once: Unified, real-time object detection.
				\newblock In {\em Proceedings of the IEEE/CVF Conference on Computer Vision and
					Pattern Recognition (CVPR)}, pages 779--788, 2016.
				
				\bibitem{redmon2017yolo9000}
				Joseph Redmon and Ali Farhadi.
				\newblock {YOLO9000}: better, faster, stronger.
				\newblock In {\em Proceedings of the IEEE/CVF Conference on Computer Vision and
					Pattern Recognition (CVPR)}, pages 7263--7271, 2017.
				
				\bibitem{redmon2018yolov3}
				Joseph Redmon and Ali Farhadi.
				\newblock {YOLOv3}: An incremental improvement.
				\newblock {\em arXiv preprint arXiv:1804.02767}, 2018.
				
				\bibitem{rezatofighi2019generalized}
				Hamid Rezatofighi, Nathan Tsoi, JunYoung Gwak, Amir Sadeghian, Ian Reid, and
				Silvio Savarese.
				\newblock Generalized intersection over union: A metric and a loss for bounding
				box regression.
				\newblock In {\em Proceedings of the IEEE/CVF Conference on Computer Vision and
					Pattern Recognition (CVPR)}, pages 658--666, 2019.
				
				\bibitem{roh2021sparse}
				Byungseok Roh, JaeWoong Shin, Wuhyun Shin, and Saehoon Kim.
				\newblock {Sparse DETR}: Efficient end-to-end object detection with learnable
				sparsity.
				\newblock {\em arXiv preprint arXiv:2111.14330}, 2021.
				
				\bibitem{sandler2018mobilenetv2}
				Mark Sandler, Andrew Howard, Menglong Zhu, Andrey Zhmoginov, and Liang-Chieh
				Chen.
				\newblock {MobileNetV2}: Inverted residuals and linear bottlenecks.
				\newblock In {\em Proceedings of the IEEE/CVF Conference on Computer Vision and
					Pattern Recognition (CVPR)}, pages 4510--4520, 2018.
				
				\bibitem{shen2019object}
				Zhiqiang Shen, Zhuang Liu, Jianguo Li, Yu-Gang Jiang, Yurong Chen, and
				Xiangyang Xue.
				\newblock Object detection from scratch with deep supervision.
				\newblock {\em IEEE Transactions on Pattern Analysis and Machine Intelligence
					(TPAMI)}, 42(2):398--412, 2019.
				
				\bibitem{simonyan2014very}
				Karen Simonyan and Andrew Zisserman.
				\newblock Very deep convolutional networks for large-scale image recognition.
				\newblock {\em arXiv preprint arXiv:1409.1556}, 2014.
				
				\bibitem{sun2021sparse}
				Peize Sun, Rufeng Zhang, Yi Jiang, Tao Kong, Chenfeng Xu, Wei Zhan, Masayoshi
				Tomizuka, Lei Li, Zehuan Yuan, Changhu Wang, et~al.
				\newblock {Sparse R-CNN}: End-to-end object detection with learnable proposals.
				\newblock In {\em Proceedings of the IEEE/CVF Conference on Computer Vision and
					Pattern Recognition (CVPR)}, pages 14454--14463, 2021.
				
				\bibitem{szegedy2015going}
				Christian Szegedy, Wei Liu, Yangqing Jia, Pierre Sermanet, Scott Reed, Dragomir
				Anguelov, Dumitru Erhan, Vincent Vanhoucke, and Andrew Rabinovich.
				\newblock Going deeper with convolutions.
				\newblock In {\em Proceedings of the IEEE/CVF Conference on Computer Vision and
					Pattern Recognition (CVPR)}, pages 1--9, 2015.
				
				\bibitem{szegedy2016rethinking}
				Christian Szegedy, Vincent Vanhoucke, Sergey Ioffe, Jon Shlens, and Zbigniew
				Wojna.
				\newblock Rethinking the inception architecture for computer vision.
				\newblock In {\em Proceedings of the IEEE/CVF Conference on Computer Vision and
					Pattern Recognition (CVPR)}, pages 2818--2826, 2016.
				
				\bibitem{tan2019efficientnet}
				Mingxing Tan and Quoc Le.
				\newblock {EfficientNet}: Rethinking model scaling for convolutional neural
				networks.
				\newblock In {\em International Conference on Machine Learning (ICML)}, pages
				6105--6114, 2019.
				
				\bibitem{tan2021efficientnetv2}
				Mingxing Tan and Quoc Le.
				\newblock {EfficientNetv2}: Smaller models and faster training.
				\newblock In {\em International Conference on Machine Learning (ICML)}, pages
				10096--10106, 2021.
				
				\bibitem{tan2020efficientdet}
				Mingxing Tan, Ruoming Pang, and Quoc~V Le.
				\newblock {EfficientDet}: Scalable and efficient object detection.
				\newblock In {\em Proceedings of the IEEE/CVF Conference on Computer Vision and
					Pattern Recognition (CVPR)}, pages 10781--10790, 2020.
				
				\bibitem{tarvainen2017mean}
				Antti Tarvainen and Harri Valpola.
				\newblock Mean teachers are better role models: Weight-averaged consistency
				targets improve semi-supervised deep learning results.
				\newblock {\em Advances in Neural Information Processing Systems (NeurIPS)},
				30, 2017.
				
				\bibitem{tian2019fcos}
				Zhi Tian, Chunhua Shen, Hao Chen, and Tong He.
				\newblock {FCOS}: Fully convolutional one-stage object detection.
				\newblock In {\em Proceedings of the IEEE/CVF International Conference on
					Computer Vision (ICCV)}, pages 9627--9636, 2019.
				
				\bibitem{tian2022fcos}
				Zhi Tian, Chunhua Shen, Hao Chen, and Tong He.
				\newblock {FCOS}: A simple and strong anchor-free object detector.
				\newblock {\em IEEE Transactions on Pattern Analysis and Machine Intelligence
					(TPAMI)}, 44(4):1922--1933, 2022.
				
				\bibitem{vasu2022improved}
				Pavan Kumar~Anasosalu Vasu, James Gabriel, Jeff Zhu, Oncel Tuzel, and Anurag
				Ranjan.
				\newblock An improved one millisecond mobile backbone.
				\newblock {\em arXiv preprint arXiv:2206.04040}, 2022.
				
				\bibitem{wang2021scaled}
				Chien-Yao Wang, Alexey Bochkovskiy, and Hong-Yuan~Mark Liao.
				\newblock {Scaled-YOLOv4}: Scaling cross stage partial network.
				\newblock In {\em Proceedings of the IEEE/CVF Conference on Computer Vision and
					Pattern Recognition (CVPR)}, pages 13029--13038, 2021.
				
				\bibitem{wang2020cspnet}
				Chien-Yao Wang, Hong-Yuan~Mark Liao, Yueh-Hua Wu, Ping-Yang Chen, Jun-Wei
				Hsieh, and I-Hau Yeh.
				\newblock {CSPNet}: A new backbone that can enhance learning capability of
				{CNN}.
				\newblock In {\em Proceedings of the IEEE/CVF Conference on Computer Vision and
					Pattern Recognition Workshops (CVPRW)}, pages 390--391, 2020.
				
				\bibitem{wang2021you}
				Chien-Yao Wang, I-Hau Yeh, and Hong-Yuan~Mark Liao.
				\newblock You only learn one representation: Unified network for multiple
				tasks.
				\newblock {\em arXiv preprint arXiv:2105.04206}, 2021.
				
				\bibitem{wang2021end}
				Jianfeng Wang, Lin Song, Zeming Li, Hongbin Sun, Jian Sun, and Nanning Zheng.
				\newblock End-to-end object detection with fully convolutional network.
				\newblock In {\em Proceedings of the IEEE/CVF Conference on Computer Vision and
					Pattern Recognition (CVPR)}, pages 15849--15858, 2021.
				
				\bibitem{wu2021fbnetv5}
				Bichen Wu, Chaojian Li, Hang Zhang, Xiaoliang Dai, Peizhao Zhang, Matthew Yu,
				Jialiang Wang, Yingyan Lin, and Peter Vajda.
				\newblock {FBNetv5}: Neural architecture search for multiple tasks in one run.
				\newblock {\em arXiv preprint arXiv:2111.10007}, 2021.
				
				\bibitem{xiong2021mobiledets}
				Yunyang Xiong, Hanxiao Liu, Suyog Gupta, Berkin Akin, Gabriel Bender, Yongzhe
				Wang, Pieter-Jan Kindermans, Mingxing Tan, Vikas Singh, and Bo Chen.
				\newblock {MobileDets}: Searching for object detection architectures for mobile
				accelerators.
				\newblock In {\em Proceedings of the IEEE/CVF Conference on Computer Vision and
					Pattern Recognition (CVPR)}, pages 3825--3834, 2021.
				
				\bibitem{xu2022pp}
				Shangliang Xu, Xinxin Wang, Wenyu Lv, Qinyao Chang, Cheng Cui, Kaipeng Deng,
				Guanzhong Wang, Qingqing Dang, Shengyu Wei, Yuning Du, et~al.
				\newblock {PP-YOLOE}: An evolved version of {YOLO}.
				\newblock {\em arXiv preprint arXiv:2203.16250}, 2022.
				
				\bibitem{yang20213d}
				Zetong Yang, Yin Zhou, Zhifeng Chen, and Jiquan Ngiam.
				\newblock {3D-MAN}: {3D} multi-frame attention network for object detection.
				\newblock In {\em Proceedings of the IEEE/CVF Conference on Computer Vision and
					Pattern Recognition (CVPR)}, pages 1863--1872, 2021.
				
				\bibitem{yu2018deep}
				Fisher Yu, Dequan Wang, Evan Shelhamer, and Trevor Darrell.
				\newblock Deep layer aggregation.
				\newblock In {\em Proceedings of the IEEE/CVF Conference on Computer Vision and
					Pattern Recognition (CVPR)}, pages 2403--2412, 2018.
				
				\bibitem{yu2021pp}
				Guanghua Yu, Qinyao Chang, Wenyu Lv, Chang Xu, Cheng Cui, Wei Ji, Qingqing
				Dang, Kaipeng Deng, Guanzhong Wang, Yuning Du, et~al.
				\newblock {PP-PicoDet}: A better real-time object detector on mobile devices.
				\newblock {\em arXiv preprint arXiv:2111.00902}, 2021.
				
				\bibitem{zhang2022dino}
				Hao Zhang, Feng Li, Shilong Liu, Lei Zhang, Hang Su, Jun Zhu, Lionel~M Ni, and
				Heung-Yeung Shum.
				\newblock {DINO}: {DETR} with improved denoising anchor boxes for end-to-end
				object detection.
				\newblock {\em arXiv preprint arXiv:2203.03605}, 2022.
				
				\bibitem{zhang2021varifocalnet}
				Haoyang Zhang, Ying Wang, Feras Dayoub, and Niko Sunderhauf.
				\newblock {VarifocalNet}: An {IoU}-aware dense object detector.
				\newblock In {\em Proceedings of the IEEE/CVF Conference on Computer Vision and
					Pattern Recognition (CVPR)}, pages 8514--8523, 2021.
				
				\bibitem{zhang2020bridging}
				Shifeng Zhang, Cheng Chi, Yongqiang Yao, Zhen Lei, and Stan~Z Li.
				\newblock Bridging the gap between anchor-based and anchor-free detection via
				adaptive training sample selection.
				\newblock In {\em Proceedings of the IEEE/CVF Conference on Computer Vision and
					Pattern Recognition (CVPR)}, pages 9759--9768, 2020.
				
				\bibitem{zhang2018shufflenet}
				Xiangyu Zhang, Xinyu Zhou, Mengxiao Lin, and Jian Sun.
				\newblock {ShuffleNet}: An extremely efficient convolutional neural network for
				mobile devices.
				\newblock In {\em Proceedings of the IEEE/CVF Conference on Computer Vision and
					Pattern Recognition (CVPR)}, pages 6848--6856, 2018.
				
				\bibitem{zhang2021bytetrack}
				Yifu Zhang, Peize Sun, Yi Jiang, Dongdong Yu, Zehuan Yuan, Ping Luo, Wenyu Liu,
				and Xinggang Wang.
				\newblock {BYTETrack}: Multi-object tracking by associating every detection
				box.
				\newblock {\em arXiv preprint arXiv:2110.06864}, 2021.
				
				\bibitem{zhang2021fairmot}
				Yifu Zhang, Chunyu Wang, Xinggang Wang, Wenjun Zeng, and Wenyu Liu.
				\newblock {FAIRMOT}: On the fairness of detection and re-identification in
				multiple object tracking.
				\newblock {\em International Journal of Computer Vision}, 129(11):3069--3087,
				2021.
				
				\bibitem{zheng2020distance}
				Zhaohui Zheng, Ping Wang, Wei Liu, Jinze Li, Rongguang Ye, and Dongwei Ren.
				\newblock {Distance-IoU} loss: Faster and better learning for bounding box
				regression.
				\newblock In {\em Proceedings of the AAAI Conference on Artificial Intelligence
					(AAAI)}, volume~34, pages 12993--13000, 2020.
				
				\bibitem{zhou2019iou}
				Dingfu Zhou, Jin Fang, Xibin Song, Chenye Guan, Junbo Yin, Yuchao Dai, and
				Ruigang Yang.
				\newblock {IoU} loss for {2D}/{3D} object detection.
				\newblock In {\em International Conference on 3D Vision (3DV)}, pages 85--94,
				2019.
				
				\bibitem{zhou2019objects}
				Xingyi Zhou, Dequan Wang, and Philipp Kr{\"a}henb{\"u}hl.
				\newblock Objects as points.
				\newblock {\em arXiv preprint arXiv:1904.07850}, 2019.
				
				\bibitem{zhou2018unetpp}
				Zongwei Zhou, Md~Mahfuzur Rahman~Siddiquee, Nima Tajbakhsh, and Jianming Liang.
				\newblock {UNet++}: A nested {U-Net} architecture for medical image
				segmentation.
				\newblock In {\em Deep Learning in Medical Image Analysis and Multimodal
					Learning for Clinical Decision Support}, 2018.
				
				\bibitem{zhu2020autoassign}
				Benjin Zhu, Jianfeng Wang, Zhengkai Jiang, Fuhang Zong, Songtao Liu, Zeming Li,
				and Jian Sun.
				\newblock {AutoAssign}: Differentiable label assignment for dense object
				detection.
				\newblock {\em arXiv preprint arXiv:2007.03496}, 2020.
				
				\bibitem{zhu2021deformable}
				Xizhou Zhu, Weijie Su, Lewei Lu, Bin Li, Xiaogang Wang, and Jifeng Dai.
				\newblock {Deformable DETR}: Deformable transformers for end-to-end object
				detection.
				\newblock In {\em Proceedings of the International Conference on Learning
					Representations (ICLR)}, 2021.
				
			\end{thebibliography}
			
		}
\end{document}